\documentclass{article}

\usepackage{neurips_2024}

\usepackage[utf8]{inputenc} % allow utf-8 input
\usepackage[T1]{fontenc}    % use 8-bit T1 fonts
\usepackage{hyperref}       % hyperlinks
\usepackage{url}            % simple URL typesetting
\usepackage{booktabs}       % professional-quality tables
\usepackage{amsfonts}       % blackboard math symbols
\usepackage{nicefrac}       % compact symbols for 1/2, etc.
\usepackage{microtype}      % microtypography
\usepackage{xcolor}         % colors
\usepackage{graphicx}
\usepackage[ruled]{algorithm2e}
\usepackage{float}
\usepackage{multirow}
\usepackage{adjustbox}
\usepackage{wrapfig}
\usepackage{subcaption}
\usepackage{thm-restate}  % restate theorem
\usepackage[normalem]{ulem}
\usepackage{amsmath}
\usepackage{cleveref}
\usepackage{booktabs}
\usepackage{subcaption}

\usepackage{custom_style}

\SetKwComment{Comment}{// }{}

\title{DiMSUM: \underline{Di}ffusion \underline{M}amba - A \underline{S}calable and \underline{U}nified Spatial-Frequency \underline{M}ethod for Image Generation}

\newcommand{\minisection}[1]{\noindent{\textbf{#1}}}

\author{%
  David S.~Hippocampus\thanks{Use footnote for providing further information
    about author (webpage, alternative address)---\emph{not} for acknowledging
    funding agencies.} \\
  Department of Computer Science\\
  Cranberry-Lemon University\\
  Pittsburgh, PA 15213 \\
  \texttt{hippo@cs.cranberry-lemon.edu} \\
}

\begin{document}

\maketitle

\begin{abstract}
    We introduce a novel state-space architecture for diffusion models, effectively harnessing spatial and frequency information to enhance the inductive bias towards local features in input images for image generation tasks. While state-space networks, including Mamba, a revolutionary advancement in recurrent neural networks, typically scan input sequences from left to right, they face difficulties with manually-defined scanning orders, especially in the processing of visual data. Our method demonstrates that integrating wavelet transformation into Mamba enhances the local structure awareness of visual inputs by disentangling them into wavelet subbands, representing both low- and high-frequency components. These wavelet-based outputs are then processed and seamlessly fused with the original Mamba outputs through a cross-attention fusion layer, combining both spatial and frequency information to optimize the order awareness of state-space models which is essential for the details and overall quality of image generation. Besides, we introduce a globally-shared transformer to supercharge the performance of Mamba, harnessing its exceptional power to capture global relationships. Through extensive experiments on standard benchmarks, our method demonstrates state-of-the-art results, achieving faster training convergence, and delivering high-quality outputs.
\end{abstract}

\section{Introduction}
\label{sec:intro}
Diffusion models \cite{song2020score, ho2020denoising} are a trending generative model technique that has gained significant attention in machine learning and computer vision. The core idea behind diffusion models is to learn how to reverse the diffusion process by gradually transforming a simple initial distribution, like Gaussian noise, into a complex data distribution. The flexibility, robust performance, and high-quality outputs make them powerful tools for advancing the state-of-the-art in generative modeling, and large diffusion-based generators have revolutionized the field of image \cite{rombach2022high,betker2023improving}, video \cite{ho2022video,ho2022imagenvideo}, and 3D synthesis \cite{poole2022dreamfusion, wang2024prolificdreamer, voleti2024sv3d}. While diffusion models initially rely on UNet architectures, recent methods have shifted gear to build upon transformer backbones. A line of works \cite{Peebles2022DiT, gao2023masked, esser2024scaling} have shown that transformer-based diffusion models are scalable and consistently offer higher generation quality than the UNet-based counterparts. 
% PixArt model series \cite{Chen2023PixArtFT,Chen2024PIXARTFA,Chen2024PixArtWT}, whose core is diffusion transformers (DiTs), show surreal text-to-image generation quality, and 
Even the most common open-source text-to-image tool, Stable Diffusion, has switched to use transformers in their upcoming release \cite{sauer2024fast}. Hence, transformers are becoming the new backbone standard for diffusion models. The power of this structure lies in the attention mechanism for capturing richer in-context relations. However, transformers have the drawback of a costly quadratic complexity, which might hinder their feasibility for high-dimensional data.

While transformers are taking over state-of-the-art diffusion backbones, a novel technique called state-space models (SSMs) has suddenly arrived, promising a better alternative. SSMs \cite{gu2022efficiently,gu2020hippo,gu2023mamba} have revolutionized the NLP field, favoring linear time complexity and excelling at long-context modeling. This type of network bears similarities to the recurrent process of RNNs while being capable of fully operating in parallel like convolutional networks. SSM promises to surpass transformers in most tasks that prioritize compute efficiency, such as long-sequence modeling. Mamba \cite{gu2023mamba} is a special type of SSM that offers greater quality by introducing time conditioning and context dependency to the hidden states. In the context of computer vision, within a very short period, this architecture has been used to address a variety of problems, including image perception \cite{zhu2024vision,huang2024localmamba,li2024mamba}, image restoration \cite{guo2024mambair,zhen2024freqmamba}, and image generation \cite{Liu2024VMambaVS,zhu2024vision,yan2023diffusion,hu2024zigma}. In diffusion-based image generation, Diffusion State Space models (DIFFUSSM) \cite{yan2023diffusion} already surpass their transformer-based counterparts.

Though showing many advantages, Mamba still has a critical weakness when processing 2D imagery data. Like vision transformers, images are divided into patches and then mapped into tokens. Mamba processes these tokens following a specific scanning order, thus introducing an inductive bias about 2D images into the model. Specifically, this order greatly impacts the interplay between image tokens, thereby affecting model performance. This characteristic is unfavorable, particularly when transformers have no such order-dependency issue. Many vision-based Mamba studies have focused on solving this problem on proposed advanced scanning mechanisms like bi-directional \cite{zhu2024vision}, cross-scanning \cite{Liu2024VMambaVS,li2024mamba}, or 8-way zigzag \cite{hu2024zigma}. Despite improving performance, these scanning techniques still fail to capture global and long-range relations and do not fully release Mamba's potential.

% In this paper, we focus on improving Mamba-based diffusion models for computer vision, and we restrict the scope of the paper to the image generation domain. We believe the failure of previous approaches in addressing the scanning order issue is rooted in their spatial-processing-only nature. The missing long-range relations can be effectively captured in the frequency spectrum. Thus, we propose incorporating frequency scanning alongside the existing spatial scanning mechanism. While a pioneer work already looked into the combination of spatial and frequency scanning in Mamba \cite{zhen2024freqmamba}, it targeted a very narrow task of image deraining and did not conduct a comprehensive study on how to integrate those features effectively.
In this paper, we enhance Mamba-based diffusion models, specifically focusing on image generation. Previous models failed to address the scanning order issue due to their exclusive reliance on spatial processing, overlooking crucial long-range relations that manifest in the frequency spectrum. We suggest a novel approach integrating frequency scanning with the conventional spatial scanning mechanism. Although initial work in Mamba has explored this combination for a limited task of image deraining \cite{zhen2024freqmamba}, it lacked a comprehensive analysis of the effective integration of these features.

Motivated by the above observation, this paper introduces DiMSUM,
% (Diffusion Mamba - A Scalable and Unified Spatial-Frequency Method)
a novel architecture that harnesses Mamba's power to unlock diffusion models' generation capabilities. Our approach enhances sensitivity to local structures and long-range dependencies by integrating wavelet transforms and spatial information. Using a query-swapped cross-attention technique, we dynamically synergize spatial and frequency information, accelerating convergence and improving image synthesis quality. Consequently, this boosts image quality and enhances the efficiency and scalability of the training.

Additionally, we incorporate globally shared transformer blocks to address global context integration, a limitation of traditional Mamba models. The block can also be viewed as a token-mixing layer that enriches global relations among image tokens, addressing the weak inductive bias of the manually defined scanning orders in the original Mamba. Hence, DiMSUM can maintain high performance even with larger, more complex datasets. Extensive experiments show that DiMSUM achieves state-of-the-art FID scores and recall, setting a new benchmark in generative image modeling.

% Motivated by this, our method introduces a novel Mamba architecture that effectively integrates frequency information with the spatial features of input images, leading to improved image generation quality and faster training convergence.
In summary, our contributions lie three-fold: (1) A novel Mamba architecture for diffusion models that leverages both spatial and frequency features to enhance the awareness of local structures within input images, leading to better image generation. (2) We interleave globally-shared transformer block per a certain number of Mamba blocks. The transformer with strong capacity for capturing global relationships significantly boost generation result when integrating with Mamba. The transformer can also be seen as an order-invariant mixing layer that complements Mamba's loose assumptions about the order of 2D data. (3) Superior results across image generation benchmarks like ImageNet, CelebA-HQ and LSUN Church. Additionally, our method maintains comparable GFLOPs and parameters with existing diffusion architectures while offering faster training convergence.

% In summary, our contributions lie in three-folds:
% \begin{itemize}
%     \item First, we introduce a novel Mamba architecture for diffusion models that leverages both spatial and frequency features to enhance the awareness of local structures within input images, leading to better image generation. %This approach enables dynamic interaction between the global and local relationships of input features through the lens of frequency domains, enhancing position awareness and spatial neighborhood of 2D input data—areas where the original Mamba scanning falls short.
%     \item We then propose a hybrid Mamba-Transformer architecture that interleaves a globally-shared transformer block with a certain number of Mamba blocks. The transformer's strong capacity for capturing global relationships is crucial for boosting generation results when integrating with Mamba. The transformer can also be seen as an order-invariant mixing layer that complements Mamba's loose assumptions about the order of 2D data.
%     \item Our method demonstrates superior results across several image generation benchmarks.
%     % , including unconditional image generation on CelebA (256 \& 512) and LSUN Church, as well as class-conditional image generation on ImageNet1K 256. 
%     Additionally, our method maintains comparable GFLOPs and parameters with existing diffusion architectures while offering faster training convergence. 
% \end{itemize}

% \metaxas{Hi Prof. Metaxas, could you use the command metaxas to comment on the paper. Thanks }
\section{Related Work}
\subsection{State Space Models and Their Applications in Vision Tasks}

% In control engineering and system identification, state space models (SSMs) are models that use state variables to describe a system by a set of first-order differential equations. The original state space model has poor performance in deep learning applications. Recently, S4 \cite{gu2022efficiently} successfully boosted the performance of state space models by carefully initializing them with a special HiPPO \cite{gu2020hippo} matrix. Follow-up state space works \cite{gu2022efficiently, gu2023mamba, gu2022parameterization, lieber2024jamba, zamba2024} achieve outstanding performance on par with transformer ones in long-range sequence modeling. Additionally, state space models are more efficient than transformer ones since the time and space complexities are linear. Mamba \cite{gu2023mamba} has recently outperformed transformer in NLP tasks by introducing a time-varying system where a set of learnable parameters depends on the context information of the given inputs, thus emphasizing the differences of hidden states within a long sequence. This is a promising candidate architecture for replacing transformers in many deep learning domains. 
In control engineering and system identification, state space models (SSMs) are described by state variables and first-order differential equations but initially underperformed in deep learning. Recent enhancements, notably by S4 \cite{gu2022efficiently} through the use of a HiPPO \cite{gu2020hippo} initialization matrix, have significantly improved SSMs. Subsequent studies \cite{gu2022efficiently, gu2023mamba, gu2022parameterization, lieber2024jamba, zamba2024} show that SSMs can match transformers in long-range sequence modeling with the added benefit of linear time and space complexities. Notably, Mamba \cite{gu2023mamba} has advanced over transformers in NLP by using a time-varying system with context-dependent parameters, enhancing the differentiation of hidden states over long sequences. This positions Mamba as a strong alternative to transformers across various domains.

In computer vision, ViM and VMamba \cite{zhu2024vision, Liu2024VMambaVS} are the first works to introduce Mamba as a building block in discriminative tasks. Sequentially, Mamba is explored in many computer vision tasks such as medical imaging \cite{U-Mamba, liu2024swin}, point clouds \cite{zhang2024point, liang2024pointmamba}, and image generation \cite{hu2024zigma, FeiDiS2024}. Similar to vision transformers, images are divided into patches, and the patches are mapped into tokens. The tokens are then arranged as a sequence following a scanning order. In vision transformers, the scanning order does not matter since attention scores are computed between every token pair. However, SSMs take into account the order information, introducing an inductive bias about 2D images into the model. Therefore, scanning order is vital in setting vision models' performance. ViM \cite{zhu2024vision} proposed bidirectional scanning order (sweep-2) for discriminative tasks. VMamba \cite{Liu2024VMambaVS} proposed cross-scanning (sweep-4) per each Mamba building block. This cross-scan improves the model performance but costs enormous overhead. MambaND \cite{li2024mamba} reduces that cost by introducing two methods: interleaved scanning and multi-head scanning. 
Interleaved scanning, which alternates the scanning order in the sequential blocks, is simpler but gains better performance in discriminative tasks.
% While interleaved scanning alternatives the scanning order in the sequential blocks, multi-head scanning modifies each Mamba block to split the token channel into many heads where each head follows a different scanning order. Interleaved scanning is more simple but gains better performance in discriminative tasks. 
% We also observe the same behavior in generation tasks via extensive experiments. 
Recently, Zigma \cite{hu2024zigma} proposed a zigzag-8 scanning order to preserve the locality property (i.e., each token is adjacent to its next and previous tokens). The zigzag-8 scanning order shows faster convergence compared to the bidirectional one. In this paper, we show that too many scanning orders, e.g., sweep-8 and zigzag-8, may introduce redundant computations and lead to worse performance compared to sweep-4. We instead find sweep-4 offers the best performance (\Cref{subsec:scan}).

% However, too many-way scanning orders could later harm the model since the head channel or the number of blocks per scanning order is inversely proportional to the number of scanning orders, which leads to poor learning per order.\hao{this sentence is a bit confusing and the reason is hard to prove here, just solely by empirical results} 

\subsection{Diffusion Architecture}
 Diffusion models \cite{ho2020denoising, song2019generative, song2021denoising, rombach2022high} are an emerging type of generative model that requires a sequential denoising process of several to thousands of steps to sample an image from initial Gaussian noise. Notably, most of them are based on Stochastic Differential Equations (SDE) that require an accumulation of additional stochastic noise at each generation step. Alternatively, there is a line of flow matching methods \cite{lipman2023flow, liu2023flow, ma2024sit, dao2023flow} that emphasize deterministic trajectories from pure Gaussian noise to the target data distribution, favoring a straighter solution path. Recent works \cite{ma2024sit, kingma2023understanding} have proved that diffusion models and flow matching are strongly correlated and can be converted into each other. In this work, we only focus on the simple objective of flow matching for our design.

Meanwhile, most methods are originally based on Unet architecture, which utilizes convolution resblock to capture local information at multiscale resolution. The attention layer is also used, interleaving between resblock layers to capture global information. Recently, the vision transformer \cite{dosovitskiy2020vit, liu2021Swin} has emerged and largely surpassed CNNs in many tasks. For diffusion image modeling, several transformer-based architectures \cite{Peebles2022DiT, bao2022all, gao2023masked} have been recently introduced. Transformer-based architectures capture global information better than Unet ones and can generate higher-quality images. Specifically, UViT \cite{bao2022all} replaced convolution resblock layers with transformer blocks and removed downsampling/upsampling blocks. DiT \cite{Peebles2022DiT} directly replaced Unet with a vision transformer. Inspired by this, MDT \cite{gao2023masked} and MaskDiT \cite{zheng2023fast} introduced a mask latent modeling approach to better capture contextual information and enhance training efficiency. Although transformer architectures achieve better image generation, these models suffer from quadratic time and memory complexity, slowing down training and inference processes. Recently, with the birth of Flash Attention \cite{dao2022flashattention, dao2023flashattention2}, 
% the memory complexity of transformers is significantly reduced to linear time, allowing faster training/inference speed
both training and inference time of these transformer-based models are significantly reduced thanks to the
% carefully read/write design to 
reduced IO bottleneck. However, the computation time complexity remains quadratic. 
% Hence, scaling up is still a challenging problem. 
Recently, the S4 \cite{gu2022efficiently} model has been introduced to effectively deal with long-range dependency in the NLP field. Furthermore, the S4 model favors the linear complexity time and space, which is more efficient than the transformer. Among S4 class models, Mamba \cite{gu2023mamba} stands out for its high performance in capturing long-range dependency. In diffusion models, DiffuSSM \cite{yan2023diffusion} adopts S4D \cite{gu2022parameterization} as a building block for their model and achieves better FID compared to transformer counterparts. Recently, Zigma \cite{hu2024zigma} utilized Mamba for diffusion architecture, using a zigzag scanning order to preserve locality-aware scanning property. 
% However, our work shows that zigzag-8 scanning has a slower convergence and worse performance than the sweep scan approach (e.g., sweep-4, sweep-8). 
Despite showing promising results, Mamba-based diffusion models still struggle to find an optimal scanning scheme to take advantage of the 2D inductive bias from images. We find these approaches stuck in spatial processing, thus failing to incorporate global and local relations. These relations can effectively captured in frequency spectrum, thus we propose to incorporate frequency scanning alongside the existing spatial scanning mechanism.

\subsection{Frequency-based networks}
Employing frequency components extracted by Fourier, Cosine, or Wavelet transform in solving vision tasks was common in classical computer vision. Many modern works still find this practice useful in improving the performance of deep neural networks. In perception tasks, several works \cite{liu2022nommer,liu2023devil, yao2022wave} integrate frequency processing in transformer architecture. NomMer \cite{liu2022nommer} applied a discrete cosine transformer into global attention to efficiently yield synergistic context from both global and local contexts. To improve Masked Image Modeling, Ge-AE \cite{liu2023devil} introduces an additional frequency decoder using Fourier transform to reconstruct the high-frequency information better. Wave-Vit \cite{yao2022wave} applies wavelet into self-attention modules to reduce the time and space complexity of the transformer architecture while still preserving the performance. Recent work Simba \cite{patro2024simba} replaced Mamba for attention in transformer block to efficient token mixing. For stable training of the mamba architecture, Fourier-based \ EinFFT layers are designed to replace MLP layers for better channel mixing. To solve the image denoising problem, FreqMamba \cite{zhen2024freqmamba} applied a wavelet and Fourier transformer to process features injected into the Mamba block. In generative modeling, several works \cite{yang2022wavegan, phung2023wavediff, yuan2023spatial} corporate wavelet frequencies into generative framework. By explicitly decomposing features/images into high- and low-frequency bands through wavelet transform, the generative model can train stably and converge faster. Furthermore, the high frequencies can be learned more efficiently, leading to a sharper synthesis image. Observing the benefit of wavelet processing in generative modeling, we apply discrete wavelet transform on local features before feeding into the Mamba layers. By using cross-attention to fuse wavelet frequency features and spatial features, our method achieves significant improvement in image synthesis compared to merely spatial feature processing.

\section{Method}
This section presents DiMSUM, a novel architecture aiming for effective and high-quality image synthesis. Preliminary knowledge will be provided in \cref{sec:prem}, then overview structure and mechanism of the proposed network (\cref{sec:overview}), and finally its core components (\cref{sec:DIM_block,sec:transformer_block}).
% This section presents our proposed DiMSUM architecture, aiming for effective and high-quality image synthesis. We first provide preliminary knowledge on the foundation techniques (\cref{sec:prem}). We then provide the overview structure and mechanism of the proposed network (\cref{sec:overview}), before diving into its core components (\cref{sec:DIM_block} and \ref{sec:transformer_block}).

\subsection{Preliminary}\label{sec:prem}
\minisection{State Space Model (SSM).} SSM is a new type of sequence model that uses an implicit hidden state $h(t) \in \bbR^{N\times L}$ to map the $1D$ input signal $x(t) \in \bbR^L$ to its corresponding output signal $y(t) \in \bbR^L$. This process is formulated by a parameter matrix $\bA \in \bbR^{N\times N}$ and two projection parameters $\bB \in \bbR^{N\times 1}$ and $\bC \in \bbR^{1\times N}$:
\begin{equation*}
    h'(t) = \bA h(t) + \bB x(t),\quad y(t) = \bC h'(t). \\
\end{equation*}
For practical usage, the continuous parameters $(\bA, \bB)$ are discretized by a time-scale parameter $\mathbf{\Delta}$ to produce discrete parameters $(\overline\bA, \overline\bB)$, following a zero-order hold (ZOH) rule:
\begin{equation*}
    \overline\bA = \exp(\mathbf{\Delta}\cdot\bA),\quad \overline\bB = (\mathbf{\Delta}\cdot\bA)^{-1}(\exp(\mathbf{\Delta}\cdot\bA) - \bI) \cdot \mathbf{\Delta}\cdot\bB).
\end{equation*}
Hence, the continuous system is rewritten as follows:
\begin{equation*}
    h_t = \overline\bA h_{t-1} + \overline{\bB} x_t,\quad y_t = \bC h_t. \\
\end{equation*}
Mamba then proposes a selective mechanism to enrich the dynamic interactions of different sequential states. In other words, the constant parameters $(\overline\bB, \bC, \mathbf{\Delta})$ are tuned into input-dependent parameters, enforcing the context awareness of input sequence states:
\begin{equation*}
    \overline\bB_t = \text{Linear}_N(x_t), \quad \bC_t = \text{Linear}_N(x_t), \quad \mathbf{\Delta_t = \text{Softplus}(\text{Broadcast}_L(\text{Linear}_{1}(x_t))}),
\end{equation*}
where $\text{Linear}_*(.)$ is a projection layer to $*$-dimensional vector, $\text{Softplus}(.) = \log(1 + \exp(.))$, and $\text{Broadcast}_L(.)$ means duplicating a single-value vector to $L$-dimensional vector. %However, this technique imposes a cost on hindering the parallel training process of the CNN kernel trick for SSMs. Mamba mitigated this bottleneck by introducing an associative scan algorithm, enabling efficient parallel training.

\minisection{Diffusion model.} Diffusion models \cite{sohl2015deep, ho2020denoising, song2019generative, song2020score, rombach2022high} are also known as score-based models that learn the transitional trajectories from a Gaussian noise to signals in the target domain. Most methods are based on Stochastic Differential Equation (SDE), requiring a larger number of function evaluations to generate an image. Recently, Flow matching \cite{lipman2023flow,liu2023flow,dao2023flow,ma2024sit} has proved to be a promising method that finds a deterministic mapping between input Gaussian noise and input data via solving Ordinary Differential Equation (ODE). Given an input data $x$ belonging to the modeling distribution $p(x)$ and a random noise $\epsilon \in \cN(0, \bI)$, the forward process is formulated as:
\begin{equation}
    x_t = x\alpha_t + \epsilon \sigma_t,
    \label{eq:fm_forward}
\end{equation}
where $x_t$ is the noise-added data at a time step $t \in [0, 1]$ and $(\alpha_t, \sigma_t)$ are time-dependent functions of $t$. Particularly, these functions are constrained such that $\alpha_1 = \sigma_0 = 0$ and  $\alpha_0 = \sigma_1 = 1$ to produce correct mapping between data $x$ at $t=0$ and noise $\epsilon$ at $t=1$. Specifically, \cite{lipman2023flow, liu2023flow, dao2023flow} use a simple linear function where $\alpha_t = 1 - t$ and $\sigma_t = t$. We employ Flow matching's training objective to estimate the velocity between noise $\epsilon$ and data $x$:
\begin{equation}
    \hat\theta = \argmin_\theta \bbE_{t, x_t}\left[ || x_1 - x_0 - v_\theta(x_t, c, t) ||\right],
\end{equation}
where $v_\theta$ is a velocity estimator implemented by a neural network with parameters $\theta$ and $c$ is an input condition (e.g., class or text). If no condition is used, $c$ is set to empty.

\minisection{Wavelet transformation.} 
% There are a variety of classical image compression methods, including Fourier transform \cite{brigham1988fft}, cosine transform \cite{ahmed1974dct}, and wavelet transform \cite{daubechies1990wavelet}. 
Among frequency transform techniques, wavelet transform stands out for its simplicity and efficiency. It preserves the structure of image space, with low-frequency subbands representing down-sampled approximations of the input image, while high-frequency ones emphasize local details such as vertical, horizontal, and diagonal edges. Particularly, Haar feature is the most prevalent wavelet transform, consisting of a low-pass filter $L = \frac{1}{\sqrt{2}} \begin{bmatrix} 1 & 1 \end{bmatrix}$ and a high-pass filter $H = \frac{1}{\sqrt{2}} \begin{bmatrix} -1 & 1 \end{bmatrix}$. To decompose an image $x \in \bbR^{H\times W}$, it needs to construct 4 kernels $LL^T, LH^T, HL^T, HH^T$, then applies them to the input image to extract corresponding subbands $\{x_{LL}, x_{LH}, x_{HL}, x_{HH} \; \vert \; x_* 
 \in \bbR^{H/2\times W/2}\}$, respectively. This process is called discrete wavelet transform (DWT). Notably, these filters are pairwise orthogonal, so an invertible matrix exists to map the data back to the original image space, coined as discrete inverse wavelet transform (IDWT). Given its benefits, we propose to use wavelet transform to supplement the local structure of frequency components into the process of Mamba, thus leading to enhanced image quality and training convergence, as demonstrated through our empirical experiments in \cref{sec:exp}.

\begin{figure}[t]
    \centering
    \includegraphics[width=.98\linewidth]{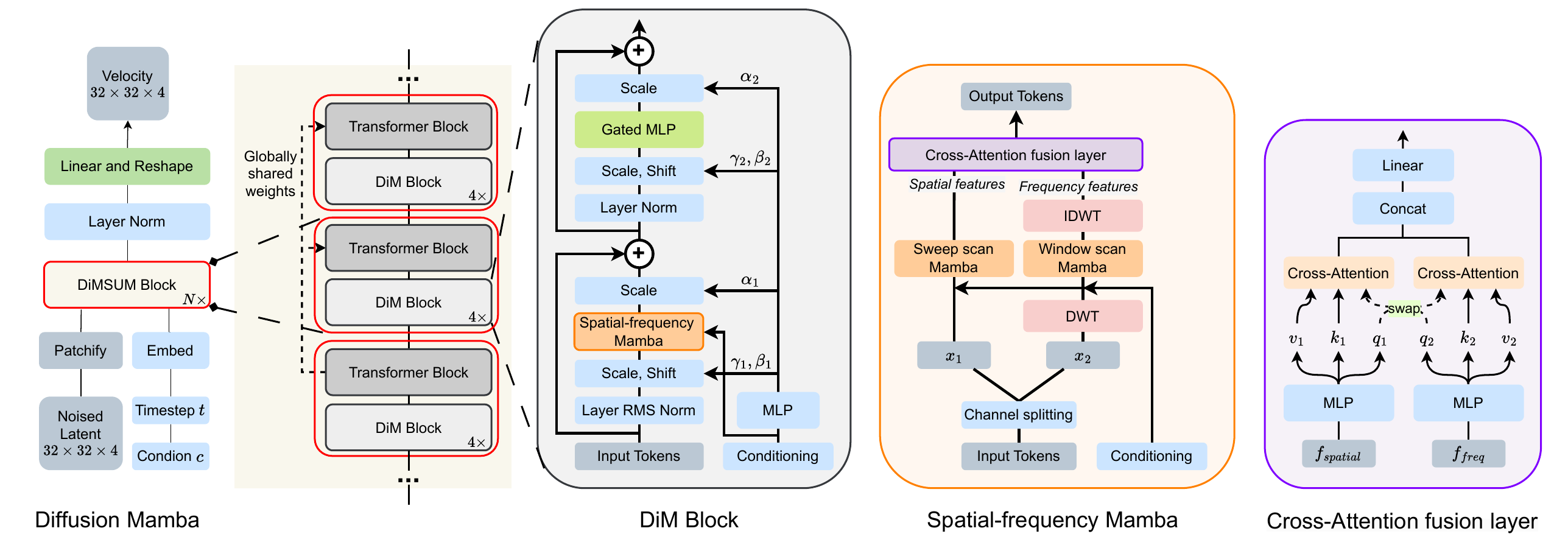}
    \vspace{-2mm}
    \caption{Overview of DiMSUM architecture.}
    \label{fig:framework}
    \vspace{-5mm}
\end{figure}

\begin{figure}[t]
    \centering
    \includegraphics[width=.8\linewidth]{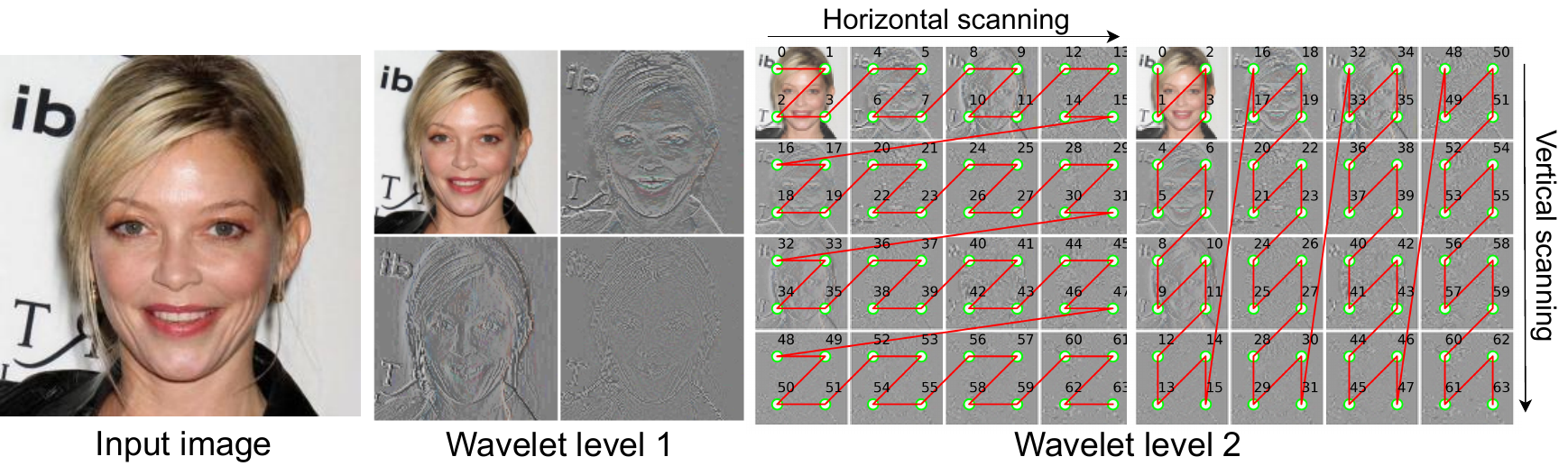}
    \vspace{-2mm}
    \caption{Illustration of Wavelet Mamba (Best view in color). For illustration purpose, we plot wavelet representations of an input image but our real process is performed on encoded features of the input. Giving an image of size $(8, 8)$, for example, it is first decomposed to four wavelet subbands of size $(4, 4)$ where each is further transformed to 2nd-level subbands of size $(2, 2)$. Green dots indicate pixel points within each wavelet subband and a window of size $2 \times 2$ is used to perform scanning across multiple wavelet subbands like the CNN kernel.}
    \label{fig:wavelet}
    \vspace{-3mm}
\end{figure}

\subsection{Overview of the proposed network}\label{sec:overview}
Inspired by the advancement of Mamba-based diffusion models and frequency-based networks, we design a novel architecture DiMSUM for effective and high-quality image synthesis with the structure presented in \Cref{fig:framework}. Similar to Latent Diffusion Models \cite{rombach2022high}, our method performs image generation on the latent space of a pre-trained encoder \footnote{https://huggingface.co/stabilityai/sd-vae-ft-ema}. The method first receives an input image and encodes it to a latent map of size $4\times H \times W$. It then processes the latent map using our proposed Diffusion Mamba network, whose core is a sequence of DiMSUM blocks, each consisting of DiM blocks that employ a novel Mamba structure with spatial and frequency scanning fusion and a globally weight-shared transformer block. The processed latent is then decoded to the output image. %In , we present an overview of our proposed network.

% \anh{In the follow-up sections, we first provide preliminary knowledge on the foundation techniques (\cref{sec:prem}). We then describe our proposed DiM block with the core is Spatial-Frequency Mamba fusion (\cref{sec:DIM_block}). Finally, we discuss the use of globally weight-shared transformer blocks as complementary components to boost up the image generation performance (\cref{sec:transformer_block})}.

\subsection{DiM block}\label{sec:DIM_block}
A core component of our approach is the DiM block that relies on a novel Spatial-Frequency Mamba fusion technique. In this section, we will discuss in detail the ideas behind this vital component.

\minisection{Wavelet Mamba.} We now examine the integration of the wavelet transform into the Mamba structure shown in \Cref{fig:wavelet}. Wavelet Mamba first applies DWT to decompose input features into wavelet subbands. Our main setting uses two-level Haar wavelet to map input into low and high-frequency features. Given input feature $x \in \bbR^{C\times H \times W}$, first-level wavelet transform is applied to produce 4 wavelet subbands of size $\bbR^{C\times H/2\times W/2}$. Each subband is then further decomposed into second-level wavelet subbands of size $\bbR^{C\times H/4\times W/4}$. This feature is pivotal, as we decompose every wavelet subband to evenly separate an input image into multiple wavelet patches, unlike conventional wavelet transformations that process only LL subbands at the next level. In Wavelet Mamba, we concatenate those subbands to form a 1D sequence, apply window scanning within each subband, and slide across the sequence for feature extraction. The window scanning is inspired by a CNN kernel proposed in \cite{huang2024localmamba} with two window sliding directions: left to right and top to bottom. Note that since low-frequency subbands capture the main content of image, it should be input first. Therefore, we do not use reverse scanning orders: right to left and bottom to top. After passing through Wavelet Mamba, the output features are transformed back to input shape by using IDWT twice. With wavelet module, our model can better capture local structure of frequency information. Thus, incorporating Wavelet Mamba with Spatial Mamba can offer better performance, yielding high-quality image generation (Spatial-frequency Mamba in \Cref{fig:framework}).

\minisection{Cross-Attention fusion layer.} Given $f_s$ and $f_w$ are spatial and wavelet features obtained by sweep and window scan. We combine these features using a cross-attention fusion layer as follows:
\begin{align*}
    &q_s, k_s, v_s = \text{Linear}(f_s),  \quad \quad q_w, k_w, v_w = \text{Linear}(f_w), \\
    % &q_w, k_w, v_w = \text{Linear}(f_w), \\
    &f_{out} = \text{Linear}(\text{Concat}(\text{Attn}(q_s, k_w, v_w), \text{Attn}(q_w, k_s, v_s))).
\end{align*}
More specifically, we first compute each feature's query ($q_*$), key ($k_*$), and value ($v_*$) using linear layers. To fuse the information between spatial and wavelet features, we do cross-attention by swapping the queries ($q_*$) of spatial and wavelet before applying a self-attention module onto each key, query, and value triplet. Finally, we concat the outputs of two cross attentions by channel followed by a linear projection to obtain the output feature $f_{out}$ (see the last subfigure in \Cref{fig:framework}).

\minisection{Conditional Mamba}. Unlike attention modules, conventional Mamba has no explicit mechanism to inject input conditions into its flow. We propose a simple technique that enables Mamba to take in any conditional input $c$ via initializing the very first hidden state with the embedding of $c$ instead of zero, as in original Mamba. This can be considered as a type of prior injection into Mamba. Specifically, the recurrent process of Mamba can be rewritten as below:
\begin{equation}
    \begin{cases}
        h_0 &= \overline\bA h_{-1} + \overline{\bB} x_0 \\
        y_0 &= \bC h_0
    \end{cases}
    ,\qquad
    \begin{cases}
        h_t &= \overline\bA h_{t-1} + \overline{\bB} x_t \\
        y_t &= \bC h_t
    \end{cases}    
\end{equation}
In conventional Mamba, $h_{-1}$ is set to zero as there is no previous state at the beginning. Here, we set $h_{-1} = \text{Linear}_D(c)$ to inject context prior into flow of Mamba. As shown in ablation, conditional mamba effectively enhances model performance. This is beneficial for both unconditional and conditional generation. For unconditional image generation, we create an auxiliary learnable token to capture image space's global information, similar to vision transformers \cite{dosovitskiy2021vit, touvron2021deit}. 
For class-conditional generation task, we use a class embedding to condition on Mamba. Conditional Mamba is enabled by default in DiM blocks (Fig. \ref{fig:framework}).

\subsection{Globally-shared attention block}\label{sec:transformer_block}
Since Mamba is better than transformer at long-range dependency \cite{gu2022efficiently, gu2023mamba} but weaker than transformer at in-context learning \cite{park2024can}, we propose a hybrid mamba-transformer architecture which favors both these properties as in recent work Jamba \cite{lieber2024jamba}. Motivated by Zamba \cite{zamba2024}, we introduce the globally-shared transformer (attention) block. This shared attention block is added after each of four DiM blocks as shown in \Cref{fig:framework} since we want to preserve the continuity of the 4-sweep alternative scanning order. By using shared weights, we significantly reduce the number of parameters introduced by different attention blocks. This layer complements the flow of Mamba since transformers excel at extracting global relations without relying on manually defined orders of input sequences, as in Mamba. Hence, with this hybrid architecture, our method effectively addresses Mamba's order dependence while significantly reducing the FID with very few additional parameters.

\section{Experiments}
\label{sec:exp}
\minisection{Implementation.} We established a depth of 20 layers, a base width of 1024, and a patch size of 2 for network configuration (further information on hyperparameters in the Appendix). We run experiments on standard datasets: CelebA-HQ\cite{celeba}, LSUN Church\cite{lsun}, and ImageNet\cite{imagenet}. For the sampling method, we follow \cite{ma2024sit, dao2023flow,hu2024zigma} to use adaptive ODE solver `dopri5' for evaluation. We assessed performance using the Fréchet Inception Distance (FID)\cite{fid} and Recall\cite{kynkaanniemi2019improved} by first generating 50K images and then comparing them with the full reference dataset. We also report GFLOPs and the number of training iterations per second to further illustrate the model efficiency.

\begin{figure}[t]
    \centering
    \begin{minipage}[c]{0.52\linewidth}
        \centering
        \begin{minipage}[c]{\linewidth}
        \centering
        \scriptsize % smaller font size
        \setlength{\tabcolsep}{3pt} % reduce padding
        \begin{tabular}{l c c c c}
            \toprule
            Model & NFE$\downarrow$ & FID$\downarrow$ & Recall$\uparrow$ & Epochs \\
            \midrule
            Ours & 61 & \textbf{4.62} & \textbf{0.52} & \textbf{225} \\
            Zigma{\textdagger} \cite{hu2024zigma} & 65 & 7.66 & 0.40 & 400 \\
            LFM-8 \cite{dao2023flow} & 89 & 5.26 & 0.46 & 500 \\
            LDM-4 \cite{rombach2022high} (ADM) & 500 & 5.11 & 0.49 & 600 \\
            LDM-8 (ADM)\textdaggerdbl & 250 & 15.37 & - & 500 \\
            LDM-8 (DiT)\textdaggerdbl & 250 & 10.21 & - & 500 \\
            LSGM \cite{vahdat2021scorebased} & 23 & 7.22 & - & 1K \\
            \midrule
            WaveDiff \cite{phung2023wavediff} & 2 & 5.94 & 0.37 & 500 \\
            DDGAN \cite{xiao2021tackling} & 2 & 7.64 & 0.36 & 800 \\
            RDUOT \cite{dao2024high}& 2 & 5.60 & 0.38 & - \\
            Score SDE \cite{song2020score} & 4000 & 7.23 & - & ~6.2K \\
            % StyleSwin \cite{zhang2022styleswin} & 1 & 3.25 & - & ~30K \\
            \bottomrule
        \end{tabular}
            \vspace{-2mm}
        \subcaption{Size $256 \times 256$. {\textdagger} is our reproduced result and {\textdaggerdbl}
 is adopted results from LFM paper.}
        \label{tab:celeb256}
        \end{minipage}
        \vspace{10mm} % Space between the table and the figure
        \begin{minipage}[c]{\linewidth}
            \centering
            \includegraphics[width=.78\linewidth]{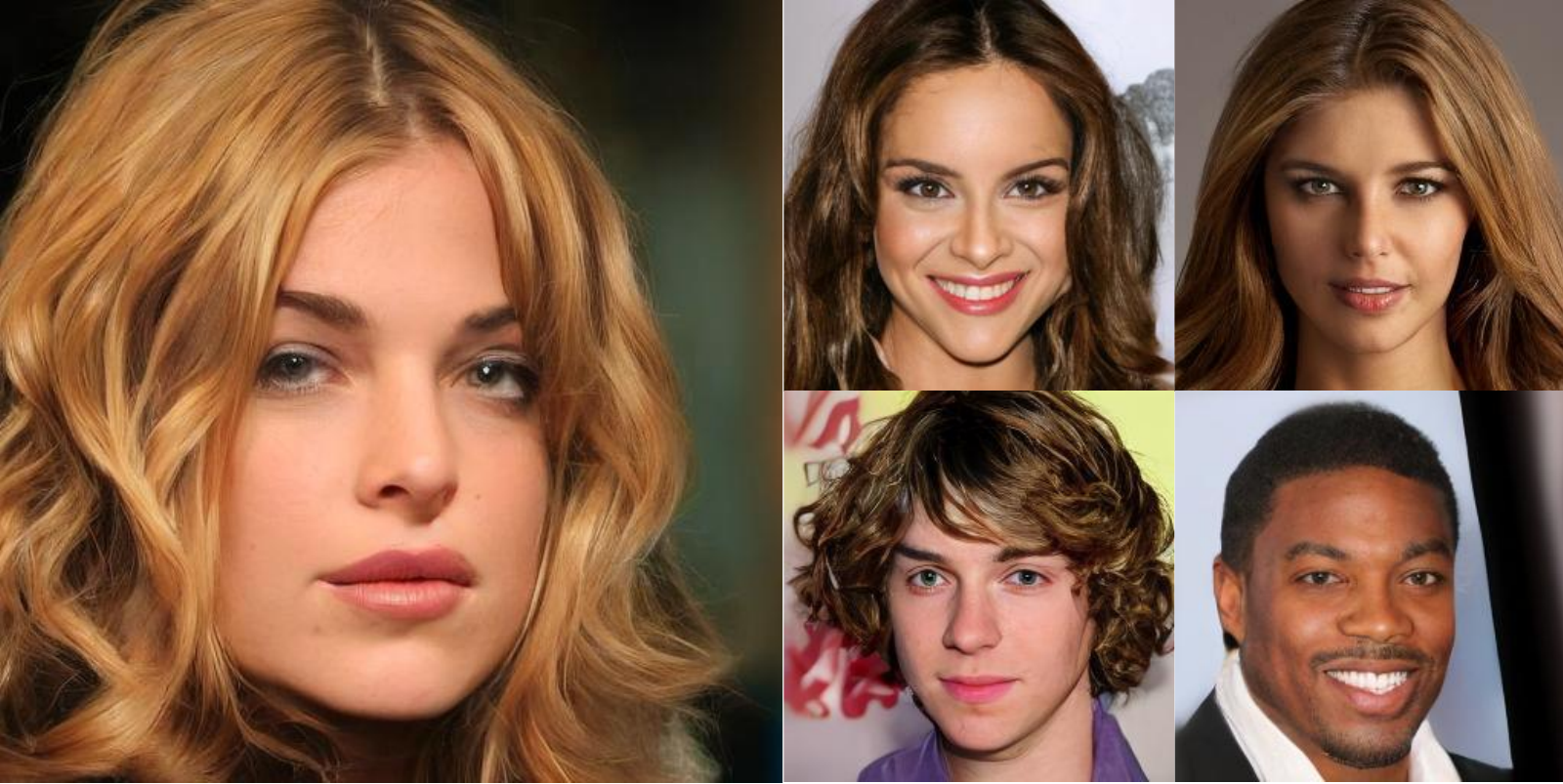}
            \vspace{-1mm}
            \subcaption{Qualitative results.}
            \label{fig:celeb_qualitative}
        \end{minipage}

    \end{minipage}%
    \hfill
    \begin{minipage}[c]{0.45\linewidth}
        \centering
        \begin{minipage}{\linewidth}
            \centering
            \scriptsize
            \setlength{\tabcolsep}{3pt}
            \begin{tabular}{l c c c c}
                \toprule
                Model & NFE$\downarrow$ & FID$\downarrow$ & Recall$\uparrow$ & Epochs \\
                \midrule
                Ours & 82 & \textbf{6.09} & \textbf{0.46} & \textbf{165} \\
                LFM-8 \cite{dao2023flow} & 94 & 6.35 & 0.41 & 500 \\
                \midrule
                WaveDiff \cite{phung2023wavediff} & 2 & 6.40 & 0.35 & 400 \\
                DDGAN \cite{xiao2021tackling} & 2 & 8.43 & 0.33 & 400 \\
                \bottomrule
            \end{tabular}
            \vspace{-2mm}
            \subcaption{Size $512 \times 512$.}
            \label{tab:celeb512}
        \end{minipage}
        \\
        % \vspace{5mm} % Space between the table and the figure
        \begin{minipage}[c]{\linewidth}
            \centering
            \includegraphics[width=.8\linewidth]{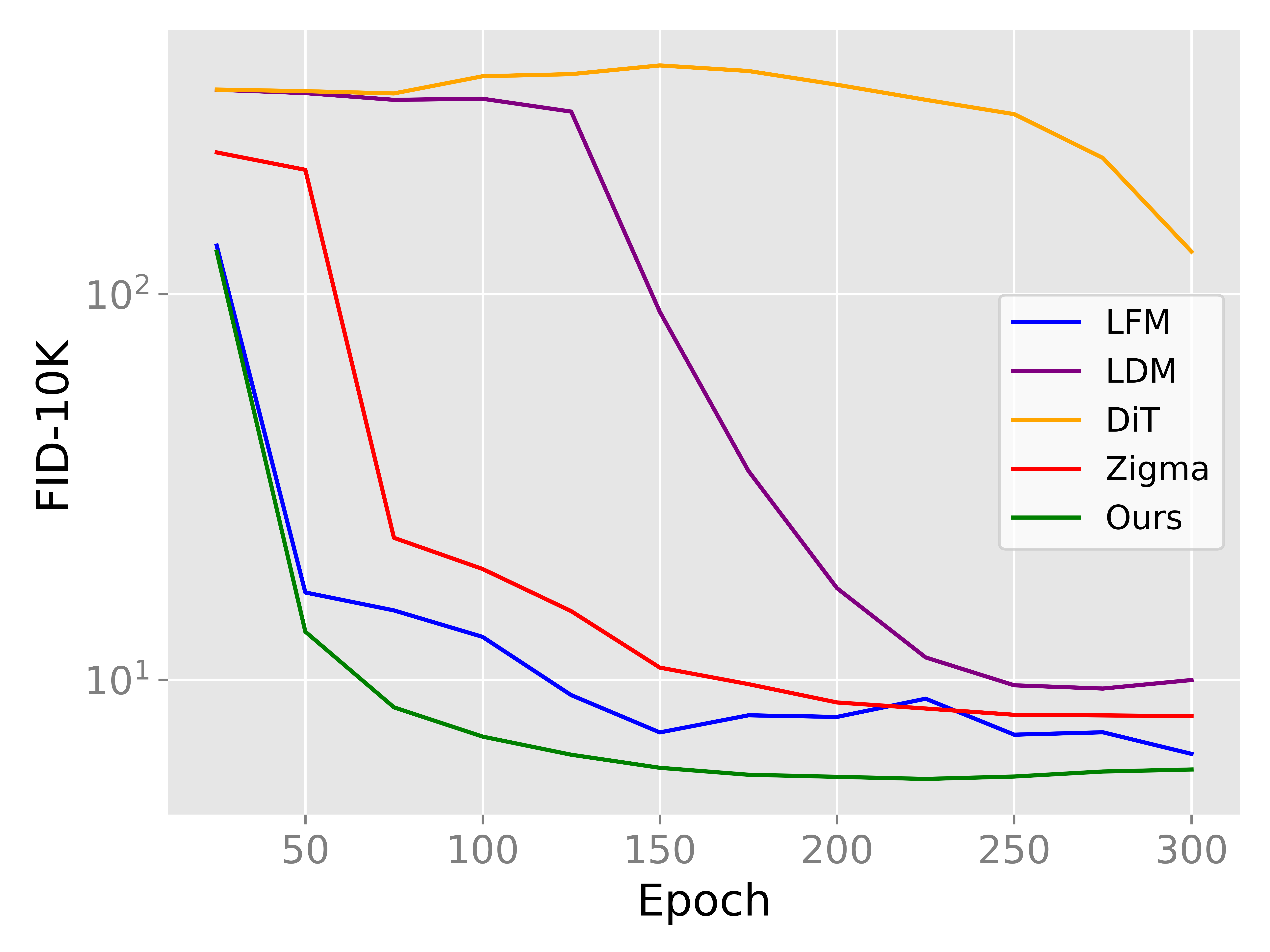}
            \vspace{-2mm}
            \subcaption{Comparison of FID-10K over training epochs.}
            \label{fig:training_convergence}
        \end{minipage}
    \end{minipage}
    \vspace{-12mm} 
    \caption{Result of our model versus others upon CelebA-HQ.  is our reproduced result based on Zigma's official code, and 
 is an adopted result from LFM paper.}
    \vspace{-2mm} 
    \label{tab:celeb}
\end{figure}

\begin{figure}[t]
    \centering
    \begin{minipage}[c]{0.6\textwidth}
        \centering
        \scriptsize
        \begin{tabular}{lcccc}
            \toprule
            Model & NFE$\downarrow$ & FID$\downarrow$ & Recall$\uparrow$ & Epochs \\
            \midrule
            Our                                 & 73  & \textbf{3.76} & \textbf{0.56} & \textbf{395} \\
            DIFFUSSM \cite{yan2024diffusion}    & 250  & 3.94          & -             & -   \\
            \midrule
            LFM-8 \cite{dao2023flow}    & 90  & 5.54          & 0.48          & 500 \\
            LDM \cite{rombach2022high}          & 250  & 4.02          & 0.52          & 400 \\
            WaveDiff \cite{phung2023wavediff}   & 2    & 5.06          & 0.40          & 500 \\
            DDPM \cite{ho2020denoising}         & 1000 & 7.89          & -             & 640 \\
            \midrule
            StyleGAN \cite{karras2019style}     & 1    & 4.21          & -             & -   \\
            StyleGAN2 \cite{karras2020training} & 1    & 3.86          & 0.36          & 13K \\
            \bottomrule
        \end{tabular}
        \subcaption{Quantitative results.}
        \label{tab:lsun256}
    \end{minipage}%
    \hfill
    \begin{minipage}[c]{0.4\textwidth}
        \centering
        \includegraphics[width=0.57\linewidth]{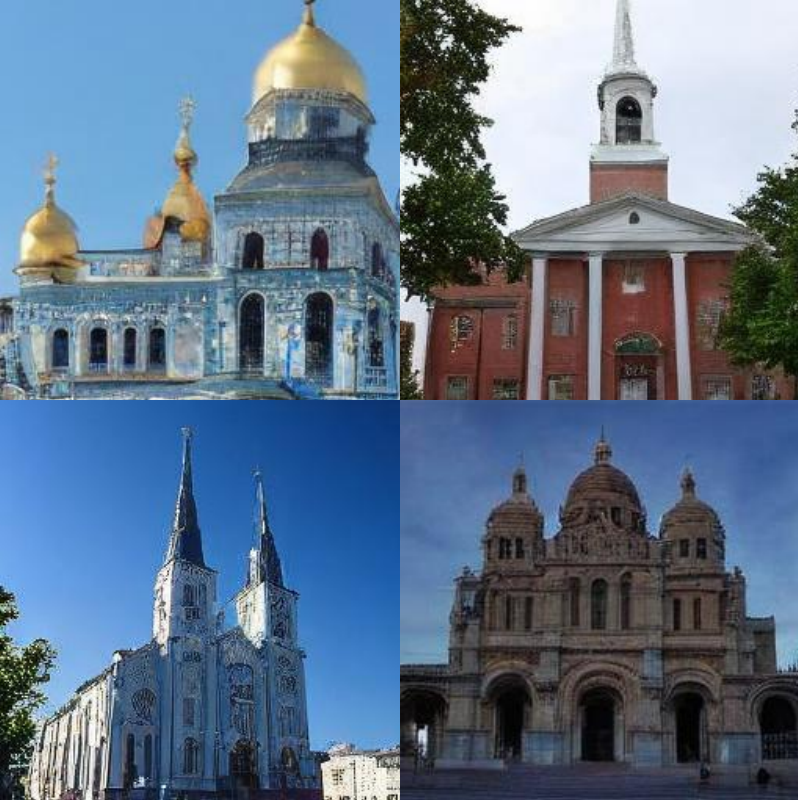}
        \subcaption{Qualitative results.}
        \label{fig:lsun_qualitative}
    \end{minipage}
    \vspace{-2mm}
    \caption{Result of our model versus others upon LSUN Church $256 \times 256$ dataset.}
    \vspace{-5mm}
\end{figure}

\subsection{Image Generation}
On the CelebA-HQ dataset at resolutions of 256x256 and 512x512 (Fig. \ref{tab:celeb}), our method achieved state-of-the-art FID scores of 4.62 and 6.09, respectively, surpassing the scores reported in recent studies. Furthermore, DiMSUM demonstrated superior recall scores, indicating a greater diversity in the generated samples compared to other methods. This result is particularly impressive, given the majority of baseline methods are based on diffusion processes, which are known for excellent diversity. On LSUN Church (Fig.  \ref{tab:lsun256}), our method outperformed diffusion-based methods and achieved results nearly on par with GAN-based approaches. Moreover, our method recorded the highest recall metric of 0.56, significantly exceeding the recall of GAN methods, thereby underscoring its robustness in generating diverse and high-quality images. 
% On ImageNet1k 256, our method achieves a score of 2.26 with guidance scale, outperforming baselines such as DIFFUSSM-XL-G, DiT, and other methods. Although our method achieves a similar score compared to SiT, it requires fewer training epochs and has a smaller parameter size.
On the ImageNet1k 256 dataset, our methodology attains a formidable FID of 2.26 using a guidance scale, surpassing the DiT models across comparable and larger model configurations, such as DiT-XL/2. This performance superiority extends to other benchmarks, including the SSM-based DIFFUSSM-XL-G model. Although our model yields results similar to those of the SiT model, our model is approximately 30\% smaller in size. %Additionally, our model demonstrates significantly accelerated convergence, evidenced by the training epoch, while continuing to deliver state-of-the-art performance.

% \begin{table}[t]
\begin{wraptable}{r}{6.5cm}
    \vspace{-12mm}
    \centering
    \caption{Class-conditional image generation on ImageNet $256 \times 256$ dataset.}
    \vspace{-2mm}
    \scriptsize
    \setlength{\tabcolsep}{4pt}
    \begin{tabular}{l c c c c}
        \toprule
    Diffusion model & FID$\downarrow$ & Recall$\uparrow$ & Params & Epochs \\
    \midrule
    % Ours (ADM) & 16.71 & 0.56 & 387M \\
    % + cfg=1.25 & 8.58 & 0.50 & 387M \\
    % Ours (DiT B/2) & 24.99 & 0.37 & 130M \\
    % + cfg=1.5 & 5.76 & 0.34 & 130M \\
    Ours &\textbf{8.61} &\textbf{0.62} & 460M & 510 \\
    Ours-G (cfg=1.5) &2.32 &0.62 & 460M & 510 \\
    Ours-G (cfg=1.4) &2.11 &0.63 & 460M & 510\\ \midrule
    
    SSM-based \\    
    % DiS-H/2 \cite{fei2024dis} (cfg=1.5) & 2.10 & 0.56 & 900M & 1.0K \\
    DIFFUSSM-XL \cite{yan2023diffusion} & 9.07 & 0.64 & 673M & 550\\
    DIFFUSSM-XL-G & 2.28 & 0.56 & 673M & 550\\
    \midrule
    % LFM (DiT B/2) & 20.38 & 0.56 & 130M & 900 \\
    % LFM (DiT B/2, cfg=1.25) & 8.77 & - & 0.48 & 130M & 900 \\
    % LFM (DiT B/2, cfg=1.5) & 4.46 & 0.42 & 130M & 900 \\
    % LDM-8 \cite{rombach2022high} & 15.51 & 0.63 & 395M & 240 \\
    % LDM-8-G & 7.76 & 0.35 & 506M & 240 \\
    UNet-based \\
    LDM-4 \cite{rombach2022high} & 10.56 & 0.62 & 400M & 200 \\
    LDM-4-G (cfg=1.5) & 3.60 & 0.48 & 400M & 200 \\
    \midrule
    % DiT-B/2 \cite{Peebles2022DiT} & 43.47 & - & - & 130M \\
    % Ours &15.29 & - &460M &80 \\
    Transformer-based \\
    DiT-L/2 \cite{Peebles2022DiT} & 23.33 & - & 458M & 80 \\
    % DiT-XL/2 & 19.47 & - & 675M & 80 \\
    DiT-XL/2 & 9.62 & 0.67 & 675M & 1.4K \\
    % DiT-XL/2-G (cfg=1.25) & 3.22 & 0.62 & 675M & 1.4K \\
    DiT-XL/2-G (cfg=1.5) & 2.27 & 0.57 & 675M & 1.4K \\
    SiT-XL/2 \cite{ma2024sit} & 9.40 & - & 675M & 1.4K \\
    SiT-XL/2-G (cfg=1.5) & 2.15 & 0.59 & 675M & 1.4K \\
    % SiT-XL/2 (SDE) & 8.60 & - & 675M & 1.4K \\
    % SiT-XL/2-G (cfg=1.5, SDE) & 2.06 & 0.59 & 675M & 1.4K \\
    
    % \midrule
    % ADM \cite{dhariwal2021diffusion} & 10.94 & 0.63 & 554M & 400 \\
    % ADM-U & 7.49 & 0.63 & 554M & 400 \\
    % ADM-G & 4.59 & 0.52 & 608M & 400 \\
    % ADM-G, ADM-U & 3.94 & 0.53 & - & 400 \\
    % \midrule
    % CDM \cite{ho2022cdm} & 4.88 & - & - & 2.2K \\
    % \midrule
    % ImageBART \cite{esser2021imagebart} & 7.44 & - & 3.5B \\
    % \midrule
    % VQGAN+T \cite{esser2021taming} & 5.88 & - & 1.3B \\
    % MaskGIT \cite{chang2022maskgit} & 6.18 & 0.51 & 227M & 280 \\
    % RQ-Transformer \cite{lee2022autoregressive} & 3.80 & - & 3.8B & - \\
    % \midrule
    \midrule
    GAN model \\
    BigGan-deep \cite{brock2018large} & 6.95 & 0.28 & 160M & - \\
    StyleGAN-XL \cite{sauer2022stylegan} & 2.30 & 0.53 & 166M & 4K \\
    % GigaGAN \cite{kang2023gigagan} & 3.45 & 0.61 & 569M & \\
    \bottomrule
    \end{tabular}
    \vspace{-1mm}
    \label{tab:full_imagenet}
% \end{table}
    \vspace{-1mm}
\end{wraptable}
\subsection{Training convergence}
% As shown in \cref{fig:training_convergence}, our method offers faster training convergence in terms of FID scores compared to other methods across training epochs. Notably, its training curve is also more stable than those of transformer-based architectures like LFM and DiT.
As reported in Tables \ref{tab:full_imagenet}, \ref{tab:celeb}, \ref{tab:lsun256}, our method requires less training epochs to reach the optimal performance compared to the baseline approach, implying a strong and fast convergence. 
To better illustrate the training convergence comparison, we illustrate in
Fig. \ref{fig:training_convergence} the performance of different diffusion-based models over training epochs regarding the FID-10K on CelebA-HQ 256. Notably, our proposed method demonstrates a superior convergence speed, rapidly decreasing FID score and stabilizing at a significantly lower value than the other methods like LFM\cite{hu2024lfm}, LDM\cite{rombach2022high}, and DiT\cite{Peebles2022DiT}. This rapid descent is particularly evident within the first 150 epochs, after which our method maintains a low FID score and still with sight on decrement, suggesting a high-quality image generation capability. Compared to the learning curves of other methods, our method exhibits a more stable trajectory without significant oscillations between training epochs. %The other methods, while also showing improvements, either stabilize at higher FID values or display a less steep decline, indicating slower convergence speeds and potentially lesser image quality at equivalent epochs.

% \begin{figure}[t]
%     \centering
%     \includegraphics[width=0.55\linewidth]{figs/training_convergence.png}
%     % \vspace{-1mm}
%     \caption{Comparison of FID-10K over training epochs on CelebA-HQ 256: Our method (red) demonstrates the fastest convergence and maintains the lowest FID score, outperforming LFM, LDM, and DiT in both speed and image quality.}
%     \label{fig:training_convergence}
%     \vspace{-3mm}
% \end{figure}

% \input{tabs/celeb256}
% \input{tabs/celeb512}

\subsection{Ablation of network design}
In this section, we ablate the design choices for our network, using experiments on the CelebA-HQ 256 dataset.
For the starting baseline, we adopt the same training settings from Zigma\cite{hu2024zigma}. We choose sweep-4 with interleave scanning order \cite{li2024mamba} by default. In Fig.\ref{tab:ab_network_comps}, with our proposed conditional Mamba, the FID score is improved from 6.19 to 5.27, and the same trend is observed for recall. Meanwhile, adding Wavelet Mamba followed by a simple concatenation layer to combine spatial and wavelet features results in a worse score of 5.87 due to the lack of strong fusion between these features. This demonstrates that our proposed cross-attention fusion layer is crucial for performance improvement, fully leveraging wavelet components to achieve a score boosted to 4.92. The performance is further enhanced to 4.66 by incorporating the weight-shared transformer block. %, which combines the best features of the Mamba and transformer architectures.

\begin{table}[t]
    \centering
    \scriptsize

    \begin{minipage}{.55\textwidth}
        \centering
    \scriptsize
    \setlength{\tabcolsep}{3pt}
    \begin{tabular}{l c c c c c}
    \toprule
     & FID$\downarrow$ & Recall$\uparrow$ & Params & GFLOPs \\
    \toprule
    Baseline & 6.19 & 0.46 & 413M & 51.65 \\
    + Conditional Mamba & 5.27 & 0.49 & 446M & 51.69 \\
    + Wavelet Mamba (w/ concat) & 5.87 & 0.47 & 394M & 56.54 \\
    + Cross-Attention fusion layer & 4.92 & 0.50 & 436M & 62.42 \\
    \rowcolor{pink!60} + Shared transformer block & 4.65 & 0.52 & 459M & 84.49 \\
    \midrule
    \end{tabular}
    \vspace{-2mm}
    \subcaption{Network components.}
    \label{tab:ab_network_comps}
    \end{minipage}\hfill
    \begin{minipage}{.45\textwidth}
        \centering
        \begin{tabular}{l c c c}
        \toprule
        \#Lv & FID$\downarrow$ & Recall$\uparrow$ & GFLOPs \\
        \midrule
        1 & 5.09 & 0.50 & 84.48 \\
        \rowcolor{pink!60} 2 & 4.65 & 0.52 & 84.49 \\
        3 & 5.23 & 0.49 & 84.50 \\
        \bottomrule
        \end{tabular}
        \subcaption{Wavelet levels.}
        \label{tab:ab_wavelet_lv}
    \end{minipage}%
    
    \begin{minipage}{.55\textwidth}
        \centering
        \begin{tabular}{l c c c c}
        \toprule
         & FID$\downarrow$ & Recall$\uparrow$ & Params & GFLOPs \\
        \toprule
        Linear & 6.00 & 0.47 & 411M & 60.83 \\
        Attention & 5.05 & 0.50 & 461M & 73.71 \\
        \rowcolor{pink!60} CAFL (swap q) & 4.92 & 0.50 & 436M & 67.27 \\
        CAFL (swap k) & 5.45 & 0.48 & 436M & 67.27 \\
        \bottomrule
        \end{tabular}
        \subcaption{Fusion layers. CAFL means Cross-Attention Fusion Layer.}
        \label{tab:ab_fusion_layer}
    \end{minipage}\hfill
    \begin{minipage}{.45\textwidth}
        \centering
        \setlength{\tabcolsep}{4pt}
        \begin{tabular}{l c c c c}
        \toprule
         & FID$\downarrow$ & Recall$\uparrow$ & Params & GFLOPs \\
        \midrule
        DCT & 5.53 & 0.50 & 436M & 67.33 \\
        EinFFT & 5.63 & 0.48 & 371M & 66.96 \\
        \rowcolor{pink!60} Wavelet & 4.92 & 0.50 & 436M & 62.42 \\
        \bottomrule
        \end{tabular}
        \subcaption{Frequency types.}
        \label{tab:ab_frequency_type}
    \end{minipage}

    % \vspace{5mm} % Provides a vertical space between the rows of tables
    \begin{minipage}{.55\textwidth}
    \centering
    \begin{tabular}{l c c c}
    \toprule
    Order & FID$\downarrow$ & Recall$\uparrow$ & iters/s $\uparrow$ \\
    \toprule
    \multicolumn{4}{c}{\textbf{Conditional Mamba Only}} \\
    \toprule
    % Uni & & & & \\
    Bi & 6.39 & 0.44 & 2.06 \\
    \rowcolor{pink!60} Sweep-4 & 5.27 & 0.49 & 2.06 \\
    Sweep-8 & 5.53 & 0.48 & 1.97 \\
    Zigzag-8 & 6.17 & 0.46 & 1.97 \\
    Jpeg-8 & 6.26 & 0.45 & 1.97 \\
    Window & 10.88 & 0.36 & 2.05 \\
    % \midrule
    % \multicolumn{5}{c}{\textbf{Wavelet Mamba only}} \\
    % \midrule
    % Sweep-4 &  &  & & \\
    % Window & 23.80 & 0.59 & 0.19 & 1.72 \\
    \midrule
    \multicolumn{4}{c}{\textbf{Spatial-frequency Mamba}} \\
    \midrule
    Sweep-4 | Sweep-4 & 5.41 & 0.49 & 1.54 \\
    \rowcolor{pink!60} Sweep-4 | Window & 4.92 & 0.50 & 1.54 \\
    \bottomrule
    \end{tabular}
    \subcaption{Scanning orders.} 
    \label{tab:ab_scanning}
    \end{minipage}\hfill
    \begin{minipage}{.45\textwidth}
        \centering
        \setlength{\tabcolsep}{4pt}
        \begin{tabular}{l c c c c}
        \toprule
         & FID$\downarrow$ & Recall$\uparrow$ & Params & GFLOPs \\
        \toprule
        \multicolumn{5}{c}{\textbf{Conditional Mamba Only}} \\
        \toprule
        GS & 5.40 & 0.49 & 469M & 78.30 \\
        \toprule
        \multicolumn{5}{c}{\textbf{Spatial-Frequency Mamba}} \\
        \toprule
        Idp & 5.08 & 0.49 & 397M & 63.37 \\
        \rowcolor{pink!60} GS & 4.65 & 0.52 & 459M & 84.49 \\
        \bottomrule
        \end{tabular}
        \subcaption{Transformer layers. GS stands for globally shared. Idp stands for Independent.}
        \label{tab:ab_transformer}
    \end{minipage}%
    \vspace{-2mm}
    \caption{Ablation studies on CelebA-HQ $256 \times 256$ dataset at epoch 250.}
    \vspace{-10mm}
\end{table}

\minisection{Design choices of fusion layer.} In Fig. \ref{tab:ab_fusion_layer}, we report metrics for different fusion layers, ranging from simple linear projection to cross-attention layers. As shown, our proposed cross-attention fusion with swapped query achieves the best results while requiring fewer parameters and GFLOPs than the attention option, with only a marginal increase in computation compared to the linear option.

\minisection{Number of wavelet levels.} 
% In Fig. \ref{tab:ab_wavelet_lv}, we evaluate the effect of wavelet levels. Increasing the number of wavelet levels will raise the required computation, such as GFLOPs, reducing the speed of training and testing.
% We find that 2 wavelet levels provide the best performance while balancing the trade-off with speed. Importantly, the choice of wavelet levels should be based on the spatial dimensions of the input features. An input image of size $256\times 256$, for instance, is encoded to a compact latent of size $32 \times 32$, which is further patchified by 2 to the small size of $16\times 16$. Hence, applying three wavelet levels results in extremely small wavelet subbands of size $2\times 2$, making it difficult to effectively leverage the wavelet components' local structure.
As shown in Fig. \ref{tab:ab_wavelet_lv}, two wavelet levels provide the best performance on the CelebA-HQ 256 dataset. We argue that the choice of wavelet levels should be based on the input resolution. An input image of size $256\times 256$, for instance, is encoded to a compact latent of size $32 \times 32$, which is further patchified by 2 to the small size of $16\times 16$. Hence, applying 3 wavelet levels results in extremely small wavelet subbands of size $2\times 2$, leading to reduced performance. %making it difficult to effectively leverage the wavelet components' local structure.

\minisection{Alternative frequency transform.} Apart from wavelet transform, we also consider different types of frequency techniques like DCT and Fourier Transform (Fig. \ref{tab:ab_frequency_type}). For DCT, we propose using the JPEG scanning strategy, as employed in JPEG compression \cite{xiao@jpeg} instead of the window scanning. 
% An illustration of the Jpeg scanning order is provided in the Appendix. 
For Fourier Features, we directly adopt the EinFFT block from SiMBA \cite{patro2024simba}. In either case, the performance drops compared with the default choice of wavelet.

\minisection{Transformer layer.} In Fig. \ref{tab:ab_transformer}, we assess the advantage of the transformer layer for both Conditional Mamba and Spatial-Frequency Mamba. As shown, the globally-shared transformer further boosts the performance of our Spatial-Frequency Mamba. In contrast, applying this layer solely to spatial Mamba increases the FID score by 0.13, highlighting the essence of our Spatial-Frequency Mamba in conjunction with the transformer layer. Meanwhile, replacing this shared layer with independent transformers 
% where shared weights are disabled for all transformer layers, 
results in a decline of 0.43 in FID and 0.03 in Recall.

\subsection{Ablation of scanning orders}
\label{subsec:scan}
In Fig. \ref{tab:ab_scanning}, we compare different scanning orders of Mamba. We keep it simple by using only Mamba block for all experiments without the globally-shared transformer and fusion layer. In Spatial-frequency Mamba, we use "Sweep-4" scanning for spatial features by default and only test other scanning methods for wavelet features. Overall, Sweep-4 performs best when combined with our proposed Conditional Mamba module. We also show that scanning with many-way orders like Sweep-8, Zigzag-8, and Jpeg-8 is not guaranteed to yield better performance than Sweep-4 scanning. In spatial-frequency Mamba, it is demonstrated that our proposed window scanning for wavelet Mamba provides a better outcome than conventional sweep-4 scanning.

\section{Conclusion}
Our paper introduces a novel, promising architecture that seamlessly integrates spatial and frequency features into Mamba process. By leveraging wavelet transform within the Mamba framework, our method enhances local structure awareness and ensures efficient spatial and frequency information fusion. This dual-focus strategy improves the detail and quality of generated images and accelerates training convergence. Our comprehensive experiments demonstrate that DiMSUM consistently outperforms state-of-the-art models of comparable size across multiple benchmarks, achieving lower FID scores and higher recall metrics, highlighting its ability to produce diverse and high-fidelity images. The proposed cross-attention fusion layer and globally shared transformer block also contribute to the model's robustness and scalability. Considering the promising results, we anticipate that future research in related domains, such as text-to-image synthesis, will adapt our backbone architecture and achieve comparable improvements in performance.

\bibliographystyle{abbrv}
\bibliography{main}

\begin{thebibliography}{10}

\bibitem{bao2022all}
F.~Bao, S.~Nie, K.~Xue, Y.~Cao, C.~Li, H.~Su, and J.~Zhu.
\newblock All are worth words: A vit backbone for diffusion models.
\newblock In {\em CVPR}, 2023.

\bibitem{betker2023improving}
J.~Betker, G.~Goh, L.~Jing, T.~Brooks, J.~Wang, L.~Li, L.~Ouyang, J.~Zhuang, J.~Lee, Y.~Guo, et~al.
\newblock Improving image generation with better captions.
\newblock {\em Computer Science. https://cdn. openai. com/papers/dall-e-3. pdf}, 2(3):8, 2023.

\bibitem{brock2018large}
A.~Brock, J.~Donahue, and K.~Simonyan.
\newblock Large scale {GAN} training for high fidelity natural image synthesis.
\newblock In {\em International Conference on Learning Representations}, 2019.

\bibitem{dao2023flow}
Q.~Dao, H.~Phung, B.~Nguyen, and A.~Tran.
\newblock Flow matching in latent space.
\newblock {\em arXiv preprint arXiv:2307.08698}, 2023.

\bibitem{dao2024high}
Q.~Dao, B.~Ta, T.~Pham, and A.~Tran.
\newblock A high-quality robust diffusion framework for corrupted dataset.
\newblock In {\em European Conference on Computer Vision}, pages 107--123. Springer, 2024.

\bibitem{dao2023flashattention2}
T.~Dao.
\newblock Flashattention-2: Faster attention with better parallelism and work partitioning.
\newblock In {\em International Conference on Learning Representations (ICLR)}, 2024.

\bibitem{dao2022flashattention}
T.~Dao, D.~Fu, S.~Ermon, A.~Rudra, and C.~R{\'e}.
\newblock Flashattention: Fast and memory-efficient exact attention with io-awareness.
\newblock {\em Advances in neural information processing systems}, 35:16344--16359, 2022.

\bibitem{dosovitskiy2020vit}
A.~Dosovitskiy, L.~Beyer, A.~Kolesnikov, D.~Weissenborn, X.~Zhai, T.~Unterthiner, M.~Dehghani, M.~Minderer, G.~Heigold, S.~Gelly, J.~Uszkoreit, and N.~Houlsby.
\newblock An image is worth 16x16 words: {Transformers} for image recognition at scale.
\newblock {\em ICLR}, 2021.

\bibitem{dosovitskiy2021vit}
A.~Dosovitskiy, L.~Beyer, A.~Kolesnikov, D.~Weissenborn, X.~Zhai, T.~Unterthiner, M.~Dehghani, M.~Minderer, G.~Heigold, S.~Gelly, J.~Uszkoreit, and N.~Houlsby.
\newblock An image is worth 16x16 words: Transformers for image recognition at scale.
\newblock In {\em International Conference on Learning Representations}, 2021.

\bibitem{esser2024scaling}
P.~Esser, S.~Kulal, A.~Blattmann, R.~Entezari, J.~M{\"u}ller, H.~Saini, Y.~Levi, D.~Lorenz, A.~Sauer, F.~Boesel, et~al.
\newblock Scaling rectified flow transformers for high-resolution image synthesis.
\newblock In {\em Forty-first international conference on machine learning}, 2024.

\bibitem{fedus2022switch}
W.~Fedus, B.~Zoph, and N.~Shazeer.
\newblock Switch transformers: Scaling to trillion parameter models with simple and efficient sparsity.
\newblock {\em Journal of Machine Learning Research}, 23(120):1--39, 2022.

\bibitem{gao2023masked}
S.~Gao, P.~Zhou, M.-M. Cheng, and S.~Yan.
\newblock Masked diffusion transformer is a strong image synthesizer.
\newblock In {\em Proceedings of the IEEE/CVF International Conference on Computer Vision (ICCV)}, 2023.

\bibitem{glorioso2024zamba}
P.~Glorioso, Q.~Anthony, Y.~Tokpanov, J.~Whittington, J.~Pilault, A.~Ibrahim, and B.~Millidge.
\newblock Zamba: A compact 7b ssm hybrid model.
\newblock {\em arXiv preprint arXiv:2405.16712}, 2024.

\bibitem{gu2024mamba}
A.~Gu and T.~Dao.
\newblock Mamba: Linear-time sequence modeling with selective state spaces.
\newblock In {\em First Conference on Language Modeling}, 2024.

\bibitem{gu2020hippo}
A.~Gu, T.~Dao, S.~Ermon, A.~Rudra, and C.~R{\'e}.
\newblock Hippo: Recurrent memory with optimal polynomial projections.
\newblock {\em Advances in neural information processing systems}, 33:1474--1487, 2020.

\bibitem{gu2022parameterization}
A.~Gu, K.~Goel, A.~Gupta, and C.~R{\'e}.
\newblock On the parameterization and initialization of diagonal state space models.
\newblock {\em Advances in Neural Information Processing Systems}, 35:35971--35983, 2022.

\bibitem{gu2022efficiently}
A.~Gu, K.~Goel, and C.~R\'e.
\newblock Efficiently modeling long sequences with structured state spaces.
\newblock In {\em The International Conference on Learning Representations ({ICLR})}, 2022.

\bibitem{gui2025depthfm}
M.~Gui, J.~S. Fischer, U.~Prestel, P.~Ma, D.~Kotovenko, O.~Grebenkova, S.~A. Baumann, V.~T. Hu, and B.~Ommer.
\newblock Depthfm: Fast monocular depth estimation with flow matching.
\newblock {\em AAAI}, 2025.

\bibitem{guo2024mambair}
H.~Guo, J.~Li, T.~Dai, Z.~Ouyang, X.~Ren, and S.-T. Xia.
\newblock Mambair: A simple baseline for image restoration with state-space model.
\newblock In {\em European conference on computer vision}, pages 222--241. Springer, 2024.

\bibitem{ho2022imagenvideo}
J.~Ho, W.~Chan, C.~Saharia, J.~Whang, R.~Gao, A.~Gritsenko, D.~P. Kingma, B.~Poole, M.~Norouzi, D.~J. Fleet, et~al.
\newblock Imagen video: High definition video generation with diffusion models.
\newblock {\em arXiv preprint arXiv:2210.02303}, 2022.

\bibitem{ho2020denoising}
J.~Ho, A.~Jain, and P.~Abbeel.
\newblock Denoising diffusion probabilistic models.
\newblock {\em Advances in Neural Information Processing Systems}, 33:6840--6851, 2020.

\bibitem{hu2024zigma}
V.~T. Hu, S.~A. Baumann, M.~Gui, O.~Grebenkova, P.~Ma, J.~Fischer, and B.~Ommer.
\newblock Zigma: A dit-style zigzag mamba diffusion model.
\newblock In {\em European Conference on Computer Vision}, pages 148--166. Springer, 2024.

\bibitem{hu2023motion}
V.~T. Hu, W.~Yin, P.~Ma, Y.~Chen, B.~Fernando, Y.~M. Asano, E.~Gavves, P.~Mettes, B.~Ommer, and C.~G. Snoek.
\newblock Motion flow matching for human motion synthesis and editing.
\newblock {\em arXiv preprint arXiv:2312.08895}, 2023.

\bibitem{celeba}
H.~Huang, z.~li, R.~He, Z.~Sun, and T.~Tan.
\newblock Introvae: Introspective variational autoencoders for photographic image synthesis.
\newblock In S.~Bengio, H.~Wallach, H.~Larochelle, K.~Grauman, N.~Cesa-Bianchi, and R.~Garnett, editors, {\em Advances in Neural Information Processing Systems}, volume~31. Curran Associates, Inc., 2018.

\bibitem{huang2024localmamba}
T.~Huang, X.~Pei, S.~You, F.~Wang, C.~Qian, and C.~Xu.
\newblock Localmamba: Visual state space model with windowed selective scan.
\newblock {\em arXiv preprint arXiv:2403.09338}, 2024.

\bibitem{kadkhodaie2023learning}
Z.~Kadkhodaie, F.~Guth, S.~Mallat, and E.~P. Simoncelli.
\newblock Learning multi-scale local conditional probability models of images.
\newblock In {\em The Eleventh International Conference on Learning Representations}, 2023.

\bibitem{karras2020training}
T.~Karras, M.~Aittala, J.~Hellsten, S.~Laine, J.~Lehtinen, and T.~Aila.
\newblock Training generative adversarial networks with limited data.
\newblock In {\em Advances in neural information processing systems}, 2020.

\bibitem{karras2019style}
T.~Karras, S.~Laine, and T.~Aila.
\newblock A style-based generator architecture for generative adversarial networks.
\newblock In {\em Proceedings of the IEEE/CVF conference on computer vision and pattern recognition}, pages 4401--4410, 2019.

\bibitem{kingma2023understanding}
D.~P. Kingma and R.~Gao.
\newblock Understanding diffusion objectives as the {ELBO} with simple data augmentation.
\newblock In {\em Thirty-seventh Conference on Neural Information Processing Systems}, 2023.

\bibitem{kynkaanniemi2019improved}
T.~Kynk{\"a}{\"a}nniemi, T.~Karras, S.~Laine, J.~Lehtinen, and T.~Aila.
\newblock Improved precision and recall metric for assessing generative models.
\newblock {\em Advances in Neural Information Processing Systems}, 32, 2019.

\bibitem{lenz2025jamba}
B.~Lenz, O.~Lieber, A.~Arazi, A.~Bergman, A.~Manevich, B.~Peleg, B.~Aviram, C.~Almagor, C.~Fridman, D.~Padnos, D.~Gissin, D.~Jannai, D.~Muhlgay, D.~Zimberg, E.~M. Gerber, E.~Dolev, E.~Krakovsky, E.~Safahi, E.~Schwartz, G.~Cohen, G.~Shachaf, H.~Rozenblum, H.~Bata, I.~Blass, I.~Magar, I.~Dalmedigos, J.~Osin, J.~Fadlon, M.~Rozman, M.~Danos, M.~Gokhman, M.~Zusman, N.~Gidron, N.~Ratner, N.~Gat, N.~Rozen, O.~Fried, O.~Leshno, O.~Antverg, O.~Abend, O.~Dagan, O.~Cohavi, R.~Alon, R.~Belson, R.~Cohen, R.~Gilad, R.~Glozman, S.~Lev, S.~Shalev-Shwartz, S.~H. Meirom, T.~Delbari, T.~Ness, T.~Asida, T.~B. Gal, T.~Braude, U.~Pumerantz, J.~Cohen, Y.~Belinkov, Y.~Globerson, Y.~P. Levy, and Y.~Shoham.
\newblock Jamba: Hybrid transformer-mamba language models.
\newblock In {\em The Thirteenth International Conference on Learning Representations}, 2025.

\bibitem{li2024mamba}
S.~Li, H.~Singh, and A.~Grover.
\newblock Mamba-nd: Selective state space modeling for multi-dimensional data.
\newblock In {\em European Conference on Computer Vision}, pages 75--92. Springer, 2024.

\bibitem{liang2024pointmamba}
D.~Liang, X.~Zhou, W.~Xu, X.~Zhu, Z.~Zou, X.~Ye, X.~Tan, and X.~Bai.
\newblock Pointmamba: A simple state space model for point cloud analysis.
\newblock In {\em The Thirty-eighth Annual Conference on Neural Information Processing Systems}, 2024.

\bibitem{lipman2023flow}
Y.~Lipman, R.~T.~Q. Chen, H.~Ben-Hamu, M.~Nickel, and M.~Le.
\newblock Flow matching for generative modeling.
\newblock In {\em The Eleventh International Conference on Learning Representations}, 2023.

\bibitem{liu2022nommer}
H.~Liu, X.~Jiang, X.~Li, Z.~Bao, D.~Jiang, and B.~Ren.
\newblock Nommer: Nominate synergistic context in vision transformer for visual recognition.
\newblock In {\em Proceedings of the IEEE/CVF Conference on Computer Vision and Pattern Recognition}, pages 12073--12082, 2022.

\bibitem{liu2023devil}
H.~Liu, X.~Jiang, X.~Li, A.~Guo, Y.~Hu, D.~Jiang, and B.~Ren.
\newblock The devil is in the frequency: Geminated gestalt autoencoder for self-supervised visual pre-training.
\newblock In {\em Proceedings of the AAAI Conference on Artificial Intelligence}, volume~37, pages 1649--1656, 2023.

\bibitem{liu2024swin}
J.~Liu, H.~Yang, H.-Y. Zhou, Y.~Xi, L.~Yu, C.~Li, Y.~Liang, G.~Shi, Y.~Yu, S.~Zhang, et~al.
\newblock Swin-umamba: Mamba-based unet with imagenet-based pretraining.
\newblock In {\em International Conference on Medical Image Computing and Computer-Assisted Intervention}, pages 615--625. Springer, 2024.

\bibitem{liu2023flow}
X.~Liu, C.~Gong, and qiang liu.
\newblock Flow straight and fast: Learning to generate and transfer data with rectified flow.
\newblock In {\em The Eleventh International Conference on Learning Representations}, 2023.

\bibitem{Liu2024VMambaVS}
Y.~Liu, Y.~Tian, Y.~Zhao, H.~Yu, L.~Xie, Y.~Wang, Q.~Ye, J.~Jiao, and Y.~Liu.
\newblock Vmamba: Visual state space model.
\newblock {\em Advances in neural information processing systems}, 37:103031--103063, 2024.

\bibitem{liu2021swin}
Z.~Liu, Y.~Lin, Y.~Cao, H.~Hu, Y.~Wei, Z.~Zhang, S.~Lin, and B.~Guo.
\newblock Swin transformer: Hierarchical vision transformer using shifted windows.
\newblock In {\em Proceedings of the IEEE/CVF international conference on computer vision}, pages 10012--10022, 2021.

\bibitem{U-Mamba}
J.~Ma, F.~Li, and B.~Wang.
\newblock U-mamba: Enhancing long-range dependency for biomedical image segmentation.
\newblock {\em arXiv preprint arXiv:2401.04722}, 2024.

\bibitem{ma2024sit}
N.~Ma, M.~Goldstein, M.~S. Albergo, N.~M. Boffi, E.~Vanden-Eijnden, and S.~Xie.
\newblock Sit: Exploring flow and diffusion-based generative models with scalable interpolant transformers.
\newblock In {\em European Conference on Computer Vision}, pages 23--40. Springer, 2024.

\bibitem{fid}
M.~F. Naeem, S.~J. Oh, Y.~Uh, Y.~Choi, and J.~Yoo.
\newblock Reliable fidelity and diversity metrics for generative models.
\newblock In {\em International conference on machine learning}, pages 7176--7185. PMLR, 2020.

\bibitem{park2024can}
J.~Park, J.~Park, Z.~Xiong, N.~Lee, J.~Cho, S.~Oymak, K.~Lee, and D.~Papailiopoulos.
\newblock Can mamba learn how to learn? a comparative study on in-context learning tasks.
\newblock {\em arXiv preprint arXiv:2402.04248}, 2024.

\bibitem{patro2024simba}
B.~N. Patro and V.~S. Agneeswaran.
\newblock Simba: Simplified mamba-based architecture for vision and multivariate time series.
\newblock {\em arXiv preprint arXiv:2403.15360}, 2024.

\bibitem{Peebles2022DiT}
W.~S. Peebles and S.~Xie.
\newblock Scalable diffusion models with transformers. 2023 ieee.
\newblock In {\em CVF International Conference on Computer Vision (ICCV)}, volume 4172, 2022.

\bibitem{phung2023wavediff}
H.~Phung, Q.~Dao, and A.~Tran.
\newblock Wavelet diffusion models are fast and scalable image generators.
\newblock In {\em Proceedings of the IEEE/CVF Conference on Computer Vision and Pattern Recognition (CVPR)}, 2023.

\bibitem{poole2023dreamfusion}
B.~Poole, A.~Jain, J.~T. Barron, and B.~Mildenhall.
\newblock Dreamfusion: Text-to-3d using 2d diffusion.
\newblock In {\em The Eleventh International Conference on Learning Representations}, 2023.

\bibitem{rombach2022high}
R.~Rombach, A.~Blattmann, D.~Lorenz, P.~Esser, and B.~Ommer.
\newblock High-resolution image synthesis with latent diffusion models.
\newblock In {\em Proceedings of the IEEE/CVF Conference on Computer Vision and Pattern Recognition}, pages 10684--10695, 2022.

\bibitem{imagenet}
O.~Russakovsky, J.~Deng, H.~Su, J.~Krause, S.~Satheesh, S.~Ma, Z.~Huang, A.~Karpathy, A.~Khosla, M.~S. Bernstein, A.~C. Berg, and L.~Fei-Fei.
\newblock Imagenet large scale visual recognition challenge.
\newblock {\em International Journal of Computer Vision}, 115:211 -- 252, 2014.

\bibitem{sauer2024fast}
A.~Sauer, F.~Boesel, T.~Dockhorn, A.~Blattmann, P.~Esser, and R.~Rombach.
\newblock Fast high-resolution image synthesis with latent adversarial diffusion distillation.
\newblock In {\em SIGGRAPH Asia 2024 Conference Papers}, pages 1--11, 2024.

\bibitem{sauer2022stylegan}
A.~Sauer, K.~Schwarz, and A.~Geiger.
\newblock Stylegan-xl: Scaling stylegan to large diverse datasets.
\newblock In {\em ACM SIGGRAPH 2022 conference proceedings}, pages 1--10, 2022.

\bibitem{schusterbauer2024boosting}
J.~Schusterbauer, M.~Gui, P.~Ma, N.~Stracke, S.~A. Baumann, V.~T. Hu, and B.~Ommer.
\newblock Boosting latent diffusion with flow matching.
\newblock In {\em ECCV}, 2024.

\bibitem{sohl2015deep}
J.~Sohl-Dickstein, E.~Weiss, N.~Maheswaranathan, and S.~Ganguli.
\newblock Deep unsupervised learning using nonequilibrium thermodynamics.
\newblock In {\em International Conference on Machine Learning}, pages 2256--2265. PMLR, 2015.

\bibitem{song2021denoising}
J.~Song, C.~Meng, and S.~Ermon.
\newblock Denoising diffusion implicit models.
\newblock In {\em International Conference on Learning Representations}, 2021.

\bibitem{song2019generative}
Y.~Song and S.~Ermon.
\newblock Generative modeling by estimating gradients of the data distribution.
\newblock {\em Advances in Neural Information Processing Systems}, 32, 2019.

\bibitem{song2020score}
Y.~Song, J.~Sohl-Dickstein, D.~P. Kingma, A.~Kumar, S.~Ermon, and B.~Poole.
\newblock Score-based generative modeling through stochastic differential equations.
\newblock In {\em International Conference on Learning Representations}, 2020.

\bibitem{touvron2021deit}
H.~Touvron, M.~Cord, M.~Douze, F.~Massa, A.~Sablayrolles, and H.~J{\'e}gou.
\newblock Training data-efficient image transformers \& distillation through attention.
\newblock In {\em International conference on machine learning}, pages 10347--10357. PMLR, 2021.

\bibitem{vahdat2021scorebased}
A.~Vahdat, K.~Kreis, and J.~Kautz.
\newblock Score-based generative modeling in latent space.
\newblock In A.~Beygelzimer, Y.~Dauphin, P.~Liang, and J.~W. Vaughan, editors, {\em Advances in Neural Information Processing Systems}, 2021.

\bibitem{voleti2024sv3d}
V.~Voleti, C.-H. Yao, M.~Boss, A.~Letts, D.~Pankratz, D.~Tochilkin, C.~Laforte, R.~Rombach, and V.~Jampani.
\newblock Sv3d: Novel multi-view synthesis and 3d generation from a single image using latent video diffusion.
\newblock In {\em European Conference on Computer Vision}, pages 439--457. Springer, 2024.

\bibitem{wang2024prolificdreamer}
Z.~Wang, C.~Lu, Y.~Wang, F.~Bao, C.~Li, H.~Su, and J.~Zhu.
\newblock Prolificdreamer: High-fidelity and diverse text-to-3d generation with variational score distillation.
\newblock {\em Advances in Neural Information Processing Systems}, 36, 2024.

\bibitem{xiao2020jpeg}
W.~Xiao, N.~Wan, A.~Hong, and X.~Chen.
\newblock A fast jpeg image compression algorithm based on dct.
\newblock In {\em 2020 IEEE International Conference on Smart Cloud (SmartCloud)}, pages 106--110, 2020.

\bibitem{xiao2021tackling}
Z.~Xiao, K.~Kreis, and A.~Vahdat.
\newblock Tackling the generative learning trilemma with denoising diffusion gans.
\newblock In {\em International Conference on Learning Representations}, 2022.

\bibitem{yan2024diffusion}
J.~N. Yan, J.~Gu, and A.~M. Rush.
\newblock Diffusion models without attention.
\newblock In {\em Proceedings of the IEEE/CVF Conference on Computer Vision and Pattern Recognition}, pages 8239--8249, 2024.

\bibitem{yang2022wavegan}
M.~Yang, Z.~Wang, Z.~Chi, and W.~Feng.
\newblock Wavegan: Frequency-aware gan for high-fidelity few-shot image generation.
\newblock In {\em European Conference on Computer Vision}, pages 1--17. Springer, 2022.

\bibitem{yao2022wave}
T.~Yao, Y.~Pan, Y.~Li, C.-W. Ngo, and T.~Mei.
\newblock Wave-vit: Unifying wavelet and transformers for visual representation learning.
\newblock In {\em European Conference on Computer Vision}, pages 328--345. Springer, 2022.

\bibitem{lsun}
F.~Yu, Y.~Zhang, S.~Song, A.~Seff, and J.~Xiao.
\newblock Lsun: Construction of a large-scale image dataset using deep learning with humans in the loop.
\newblock {\em ArXiv}, abs/1506.03365, 2015.

\bibitem{yuan2023spatial}
X.~Yuan, L.~Li, J.~Wang, Z.~Yang, K.~Lin, Z.~Liu, and L.~Wang.
\newblock Spatial-frequency u-net for denoising diffusion probabilistic models.
\newblock {\em arXiv preprint arXiv:2307.14648}, 2023.

\bibitem{zhang2024point}
T.~Zhang, H.~Yuan, L.~Qi, J.~Zhanng, Q.~Zhou, S.~Ji, S.~Yan, and X.~Li.
\newblock Point cloud mamba: Point cloud learning via state space model.
\newblock {\em AAAI}, 2025.

\bibitem{zheng2024fast}
H.~Zheng, W.~Nie, A.~Vahdat, and A.~Anandkumar.
\newblock Fast training of diffusion models with masked transformers.
\newblock {\em Transactions on Machine Learning Research}, 2024.

\bibitem{FeiDiS2024}
F.~Zhengcong, F.~Mingyuan, Y.~Changqian, and H.~Jusnshi.
\newblock Scalable diffusion models with state space backbone.
\newblock {\em arXiv preprint}, 2024.

\bibitem{zhu2024vim}
L.~Zhu, B.~Liao, Q.~Zhang, X.~Wang, W.~Liu, and X.~Wang.
\newblock Vision mamba: Efficient visual representation learning with bidirectional state space model.
\newblock In {\em Forty-first International Conference on Machine Learning}.

\bibitem{zhen2024freqmamba}
Z.~Zou, H.~Yu, J.~Huang, and F.~Zhao.
\newblock Freqmamba: Viewing mamba from a frequency perspective for image deraining.
\newblock In {\em ACM Multimedia 2024}, 2024.

\end{thebibliography}

\newpage
\appendix

\section{Training details}
\label{ap:training_details}
We provide network configuration details in \cref{tab:dimsum_config} and hyperparams in \cref{tab:hyperparams}. For the interpolent form of the forward process \cref{eq:fm_forward}, we opt to use a generalized VP form, which is proposed in SiT paper \cite{ma2024sit}. Particularly, this form defines $\alpha_t = \cos{\left(\frac{1}{2}\pi t\right)}$ and $\sigma_t = \sin{\left(\frac{1}{2} \pi t\right)}$. The choice of transformer block, we adopt a DiT block \cite{Peebles2022DiT} where we replace the original MLP layer with a Gated MLP layer, similar to SwitchTransformer \cite{fedus2022switch}. 

\begin{figure}
    \centering
    \includegraphics[width=0.75\linewidth]{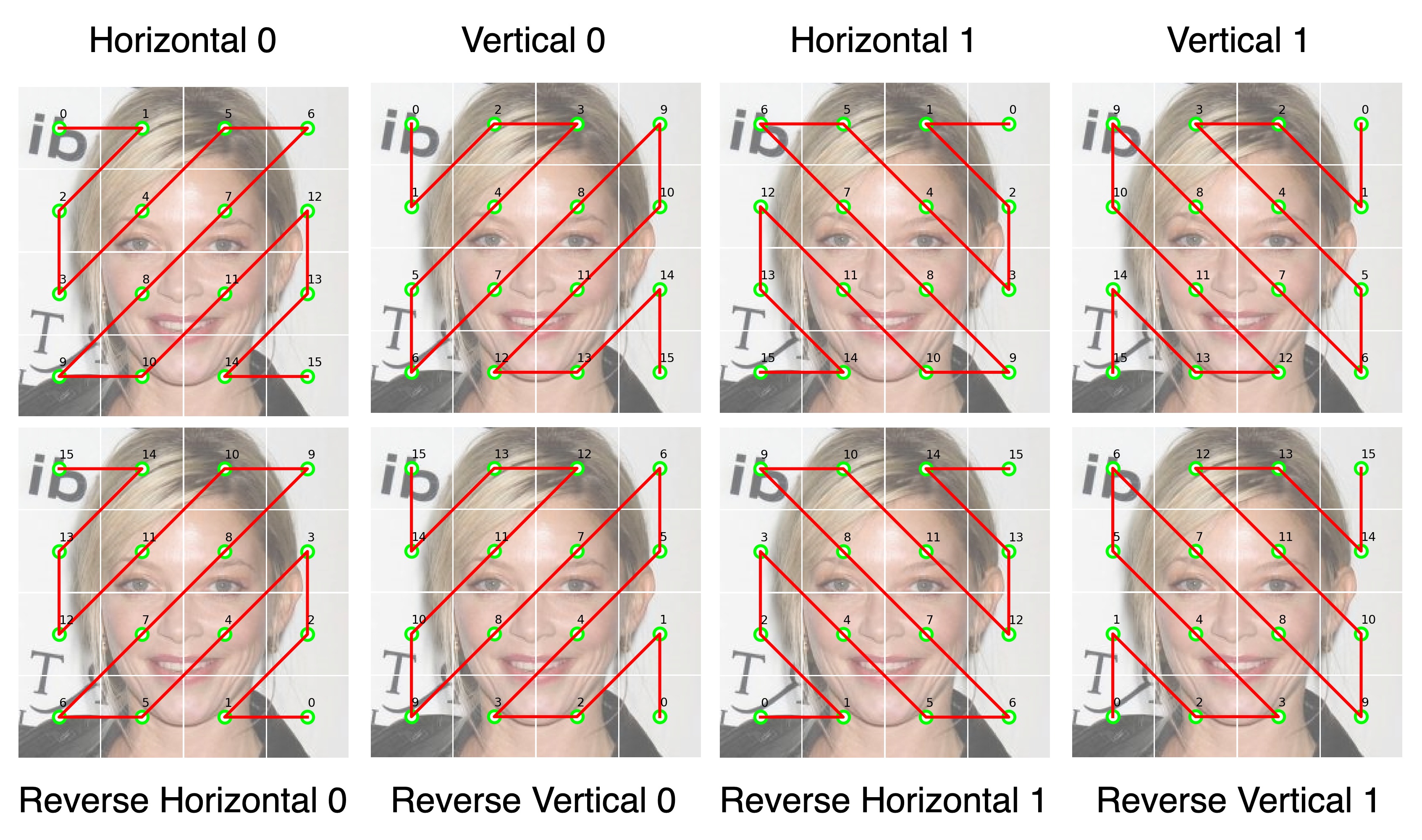}
    \caption{Illustration of Jpeg-8 scanning orders.}
    \label{fig:jpeg8}
\end{figure}

\begin{figure}
    \centering
    \includegraphics[width=0.75\linewidth]{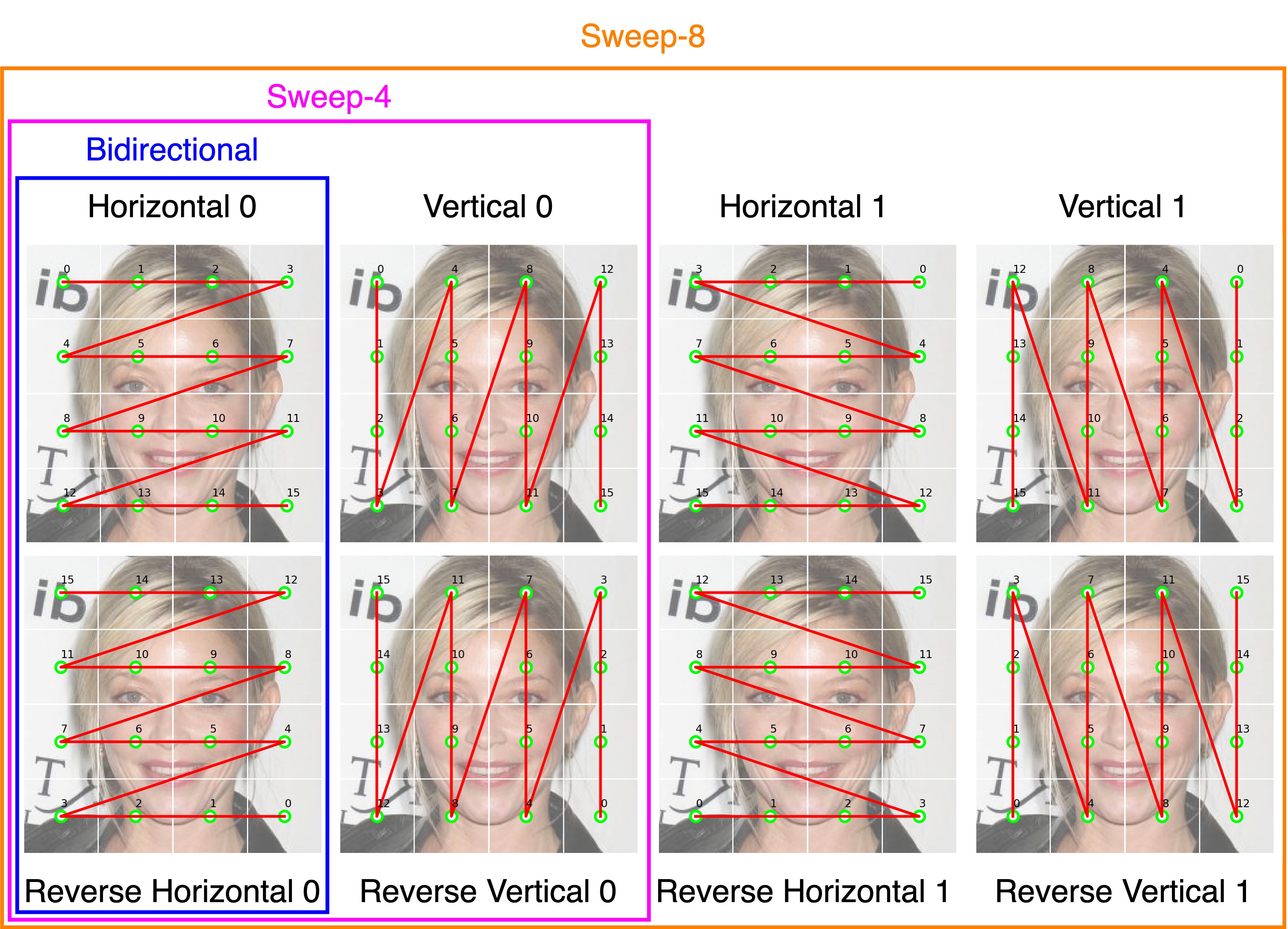}
    \caption{Illustration of Sweep scanning orders.}
    \label{fig:sweep}
\end{figure}

Besides, we illustrate the sweep scanning strategy in \Cref{fig:sweep} for better comprehension. We also visualize the Jpeg scanning orders in \Cref{fig:jpeg8} inspired by the Jpeg compression algorithm \cite{xiao@jpeg}. 

\section{More qualitative examples}
\begin{figure}
    \centering
    \includegraphics[width=1.\linewidth]{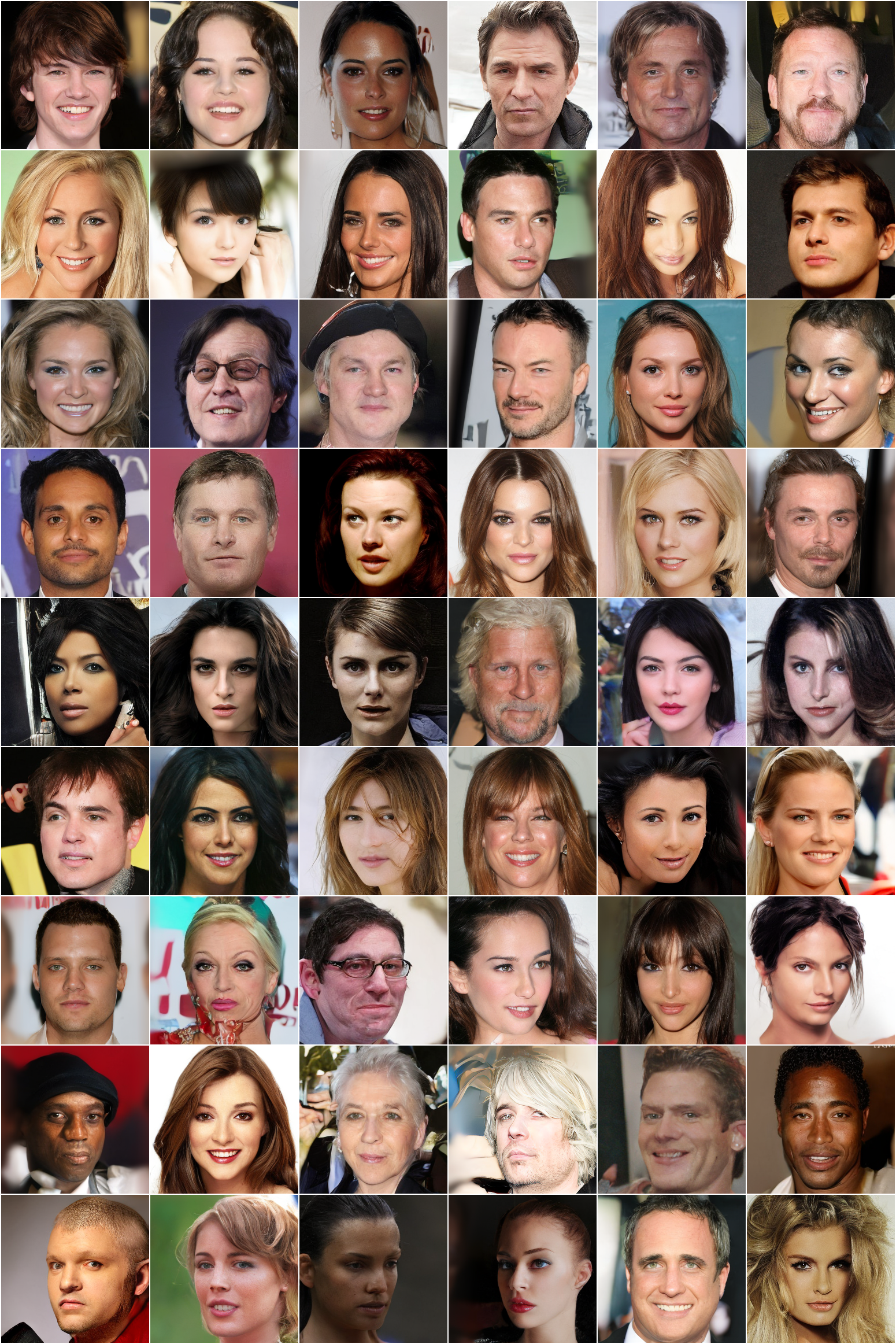}
    \caption{Uncurated generated samples of CelebA-HQ 256.}
    \label{fig:uncurated_celeb256}
\end{figure}

\begin{figure}
    \centering
    \includegraphics[width=1.\linewidth]{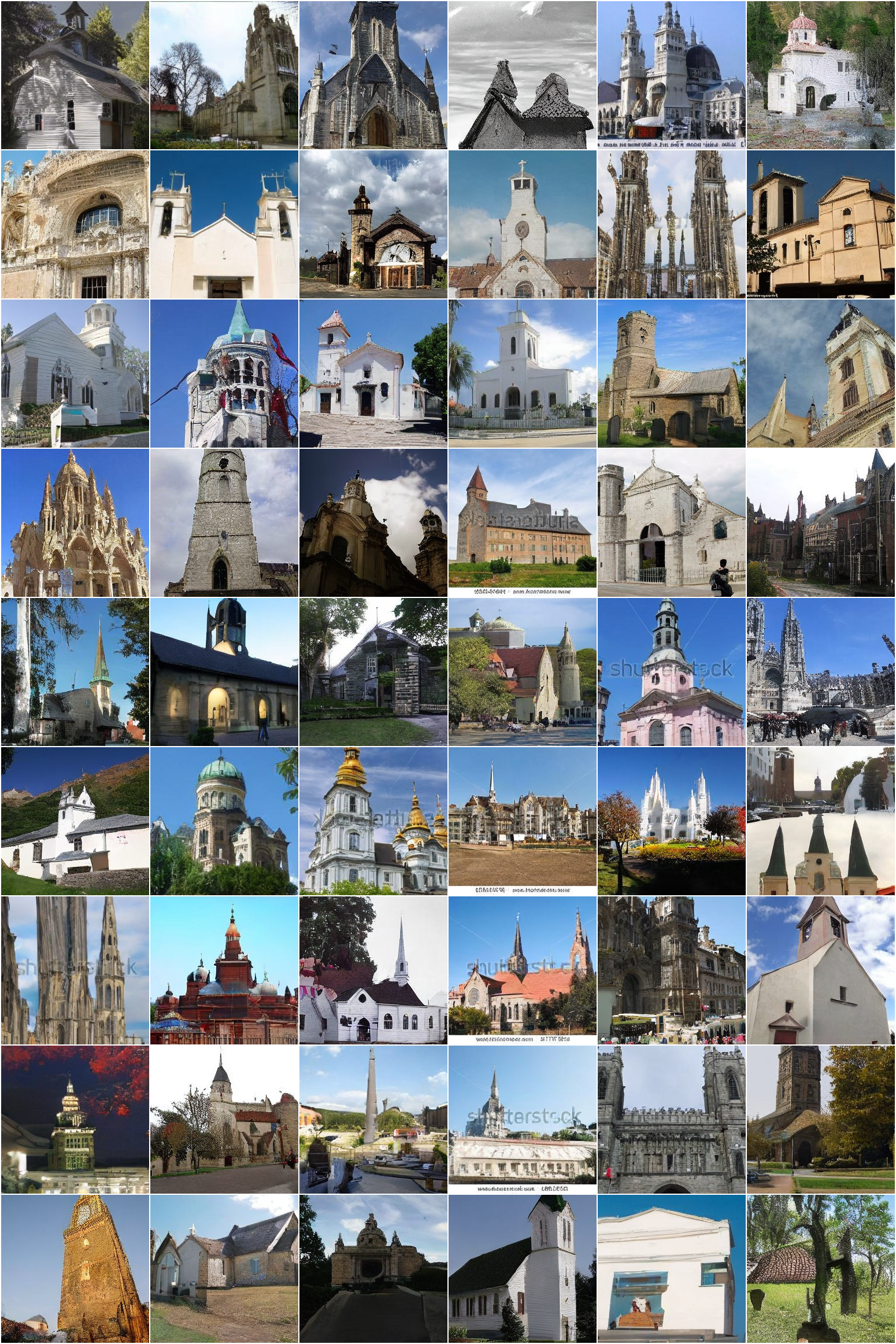}
    \caption{Uncurated generated samples of LSUN Church 256.}
    \label{fig:uncurated_church}
\end{figure}

\begin{figure}
    \centering
    \includegraphics[width=1.\linewidth]{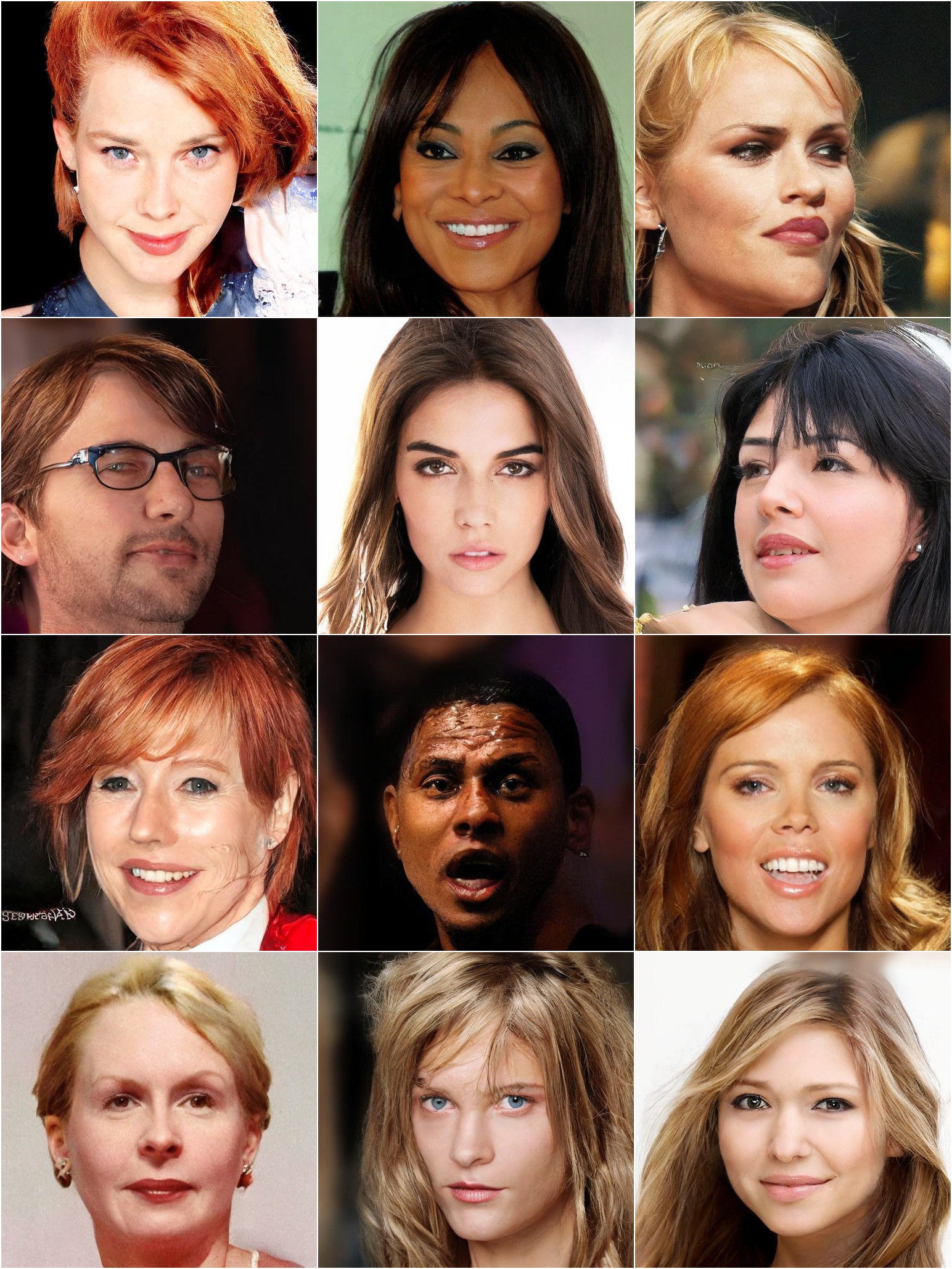}
    \caption{Uncurated generated samples of CelebA-HQ 512.}
    \label{fig:uncurated_celeb512}
\end{figure}

\begin{figure}
    \begin{minipage}[c]{0.45\linewidth}
    \includegraphics[width=\linewidth]{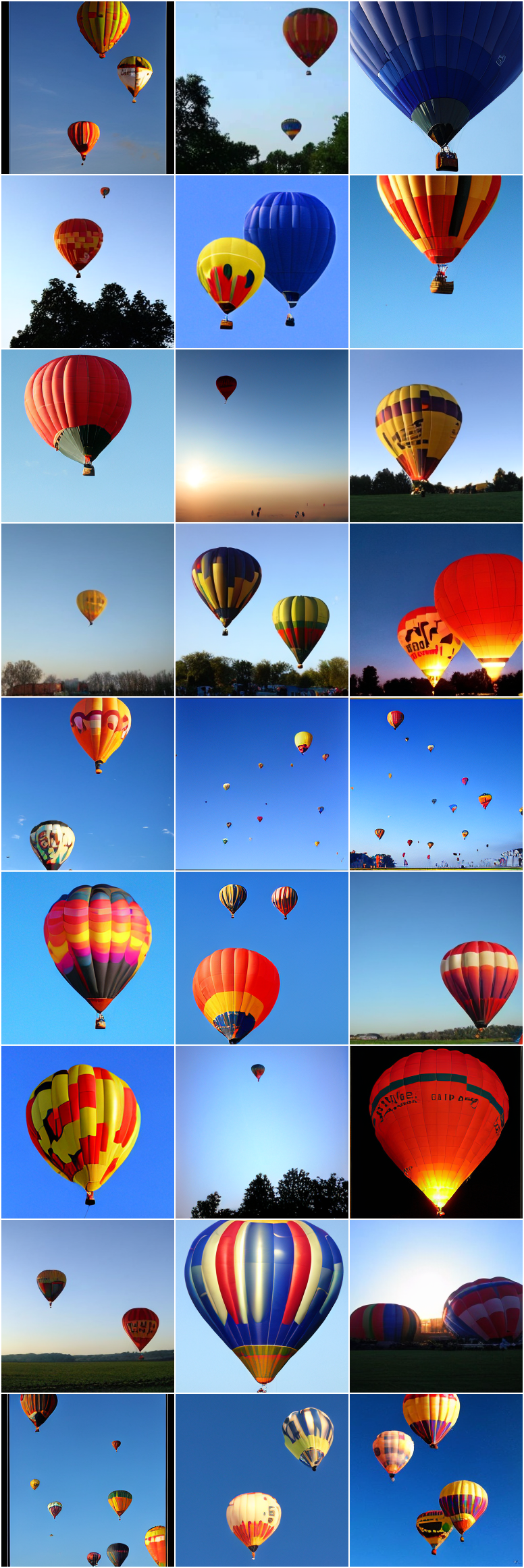}
    \caption{Uncurated generated samples of ImageNet 256 class 417 (balloon) with cfg=4.0.}
    \label{fig:imgnetsupp_2}
    \end{minipage}
    \hfill
    \begin{minipage}[c]{0.45\linewidth}
    \includegraphics[width=\linewidth]{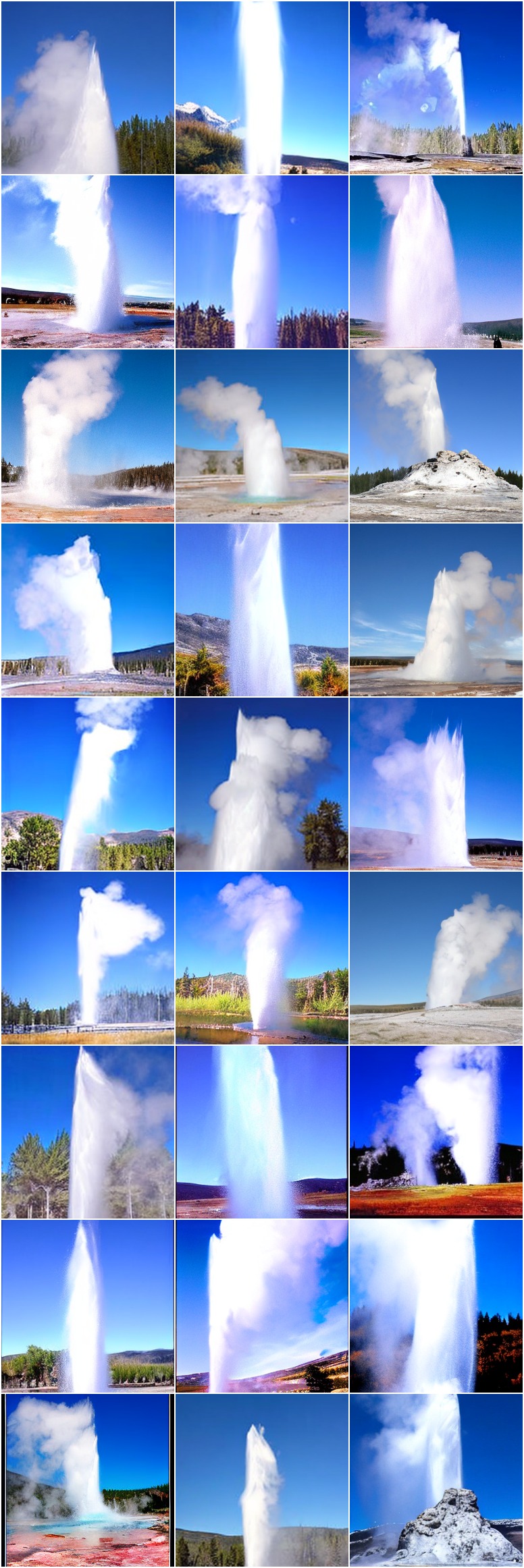}
    \caption{Uncurated generated samples of ImageNet 256 class 974 (geyser) with cfg=4.0.}
    \label{fig:imgnetsupp_3}
    \end{minipage}%
\end{figure}

\begin{figure}
    \begin{minipage}[c]{0.45\linewidth}
    \includegraphics[width=\linewidth]{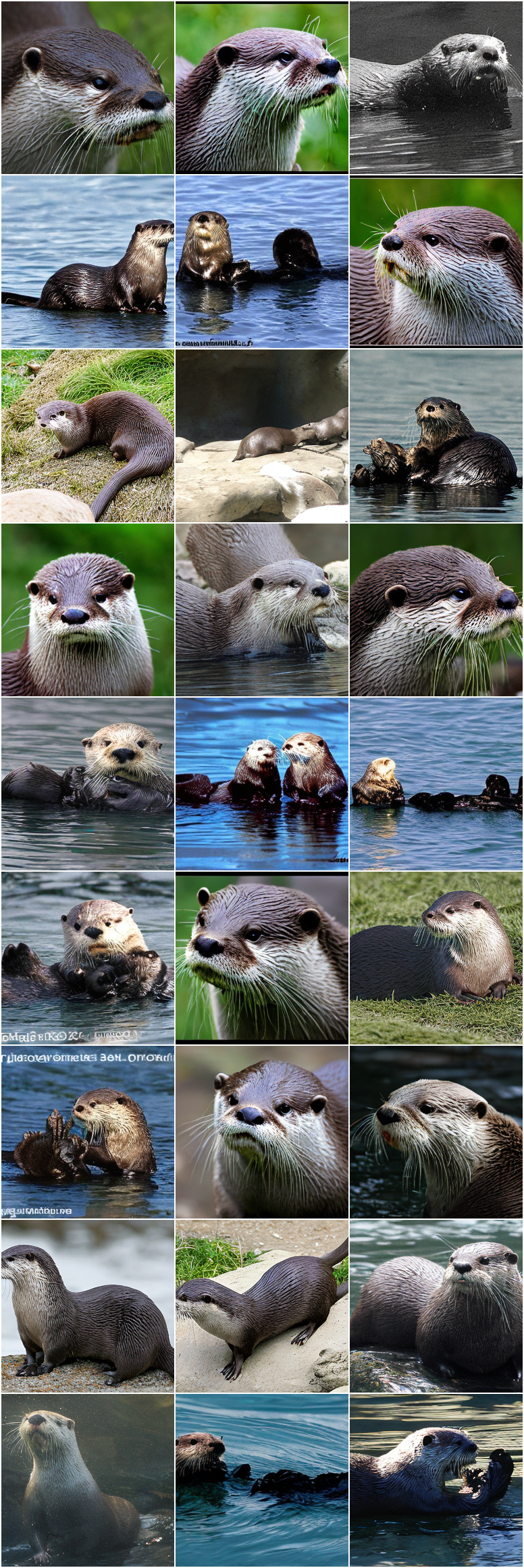}
    \caption{Uncurated generated samples of ImageNet 256 class 360 (otter) with cfg=4.0.}
    \label{fig:imgnetsupp_4}
    \end{minipage}
    \hfill
    \begin{minipage}[c]{0.45\linewidth}
    \includegraphics[width=\linewidth]{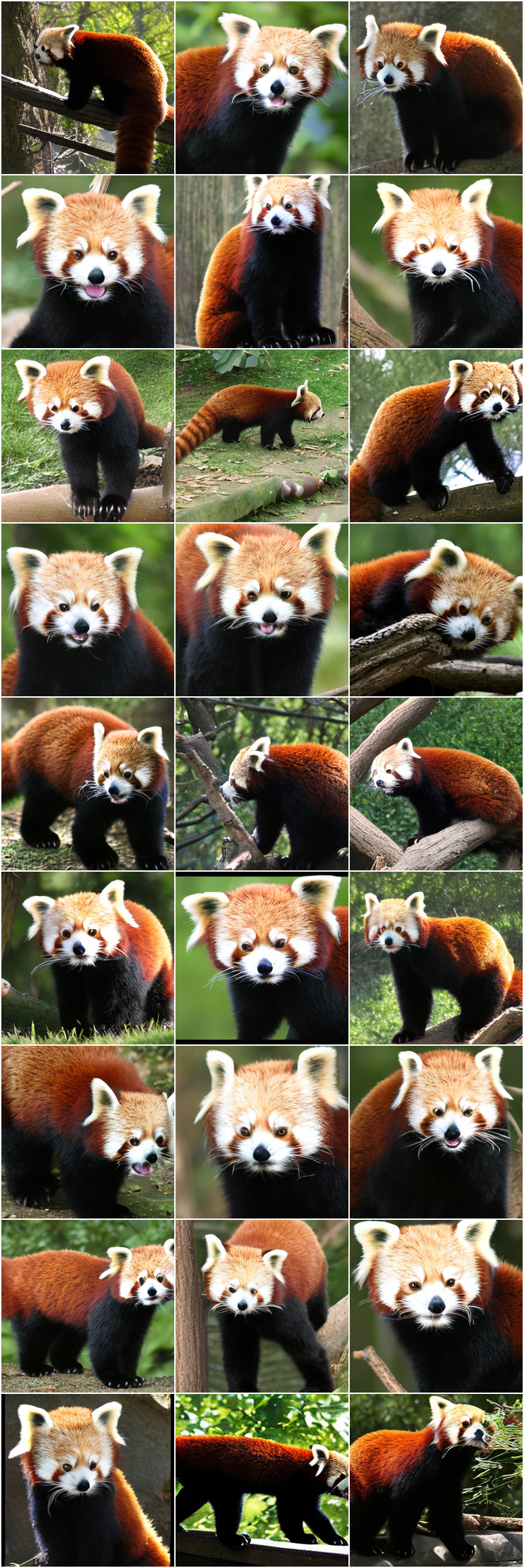}
    \caption{Uncurated generated samples of ImageNet 256 class 387 (lesser panda, red panda, panda, bear cat, cat bear, Ailurus fulgens) with cfg=4.0.}
    \label{fig:imgnetsupp_5}
    \end{minipage}%
\end{figure}

\begin{figure}
    \begin{minipage}[c]{0.45\linewidth}
    \includegraphics[width=\linewidth]{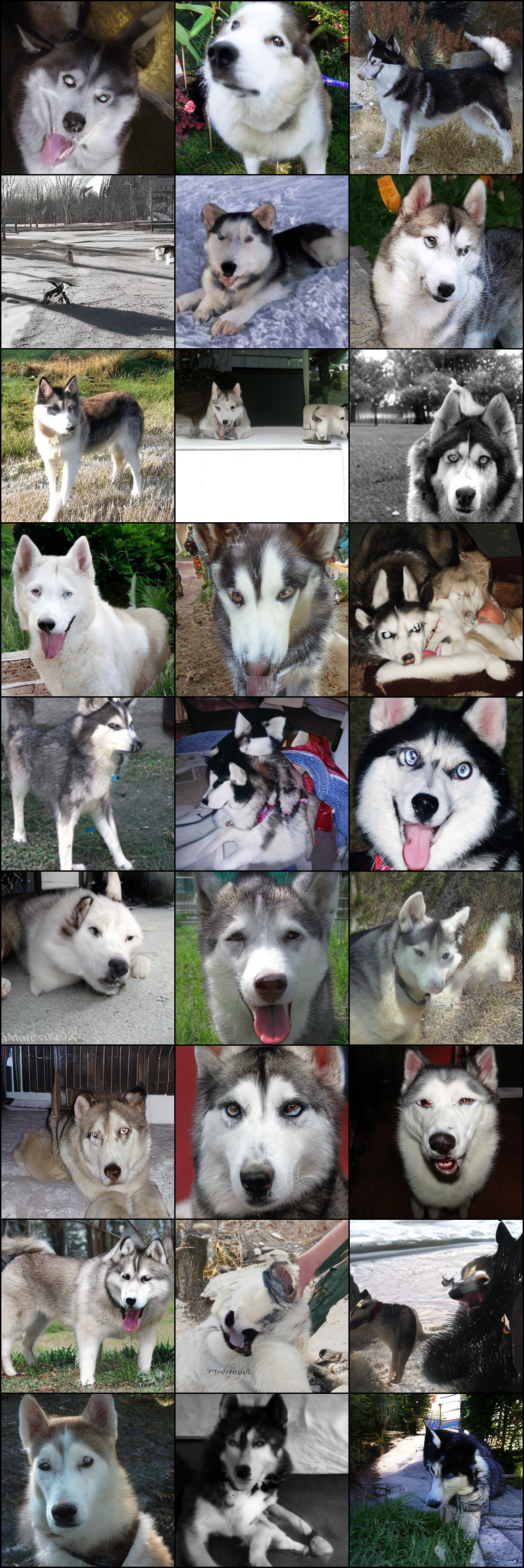}
    \caption{Uncurated generated samples of ImageNet 256 class 250 (Siberian husky) with cfg=1.5.}
    \label{fig:imgnetsupp_6}
    \end{minipage}
    \hfill
    \begin{minipage}[c]{0.45\linewidth}
    \includegraphics[width=\linewidth]{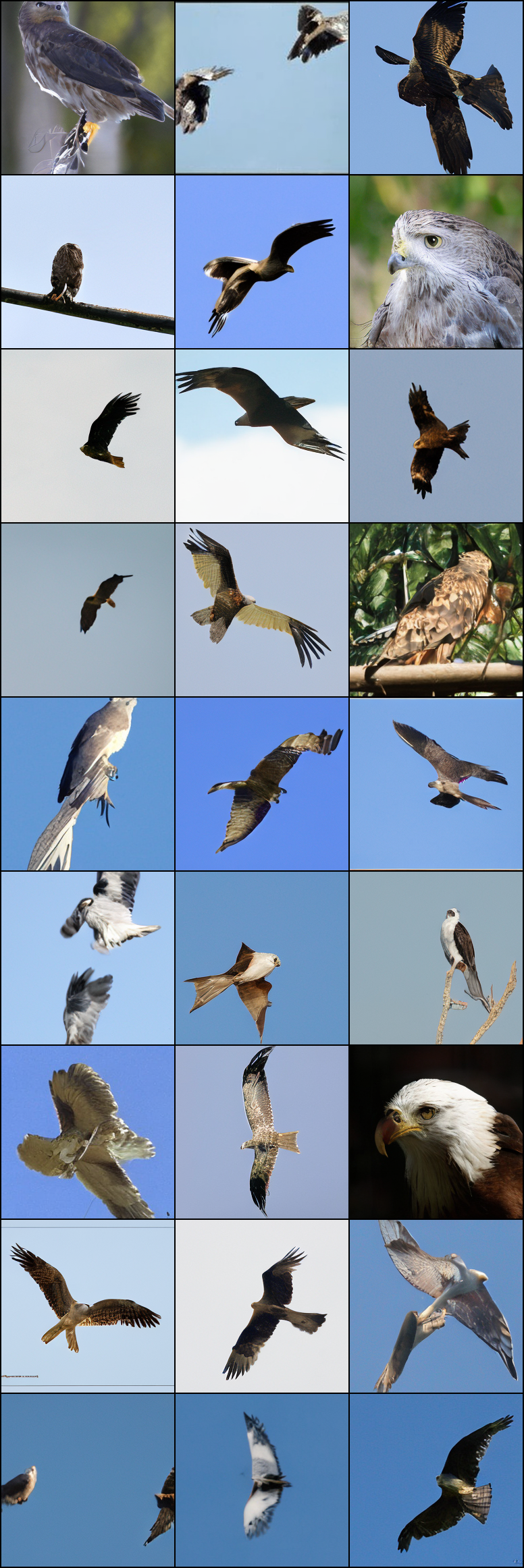}
    \caption{Uncurated generated samples of ImageNet 256 class 21 (bald eagle, American eagle, Haliaeetus leucocephalus) with cfg=1.5.}
    \label{fig:imgnetsupp_7}
    \end{minipage}%
\end{figure}

We present our uncurated generated samples of CelebA-HQ 256 in \Cref{fig:uncurated_celeb256}, CelebA-HQ 512 in \Cref{fig:uncurated_celeb512}, LSUN Church in \Cref{fig:uncurated_church}, and ImageNet in \ref{fig:imgnetsupp_0}, \ref{fig:imgnetsupp_1}, \ref{fig:imgnetsupp_2}, \ref{fig:imgnetsupp_3}, \ref{fig:imgnetsupp_4}, \ref{fig:imgnetsupp_5}, \ref{fig:imgnetsupp_6}, \ref{fig:imgnetsupp_7}.

\begin{table}[t]
    \centering
    \footnotesize
    \caption{Hyper-parameters and network config of our DiMSUM network.}
    
    \begin{subtable}{.6\linewidth}
      \centering
        \caption{Hyper-parameters}
        \begin{tabular}{lccc}
        \toprule
                      & CelebA 256 \& 512       & Church & ImageNet      \\ \midrule
        Learning rate & $1\text{e-4}$ & $5\text{e-}5$ & $1\text{e-}4$ \\
        $\beta_1,\beta_2$ & 0.9, 0.999    & 0.9, 0.999    & 0.9, 0.999    \\
        Batch size    & 64 \& 32            &  128              & 704           \\
        Droppath      & 0.1           & 0.2            & 0.1           \\
        Max-grad-norm & 2             & 2              & 1             \\
        Label-dropout & 0.            & 0.             & 0.15          \\
        Epochs        & 250 \& 165          & 395            & 320              \\
        GPUs (A100)         & 2 \& 4            &   4             & 8             \\
        Train days    & 0.89 \& 3.2        & 3.42               & 12             \\ \bottomrule
        \end{tabular}
        \label{tab:hyperparams}
    \end{subtable}%
    \begin{subtable}{.4\linewidth}
      \centering
        \caption{Network config}
        \begin{tabular}{lc}
        \toprule
        Config                  & \multicolumn{1}{l}{Value} \\ \midrule
        Depth                   & 16                        \\
        Hidden size             & 1024                      \\
        Patch size              & 2                         \\
        \begin{tabular}[c]{@{}l@{}}Use learnable absolute\\ positional embedding\end{tabular} & True \\
        Attention every k layer & 4                         \\ \bottomrule
        \end{tabular}
        \label{tab:dimsum_config}
    \end{subtable} 
\end{table}

% \begin{figure}
% \centering
% \begin{subfigure}{.45\textwidth}
%   \centering
%   \includegraphics[width=\linewidth]{figs/sit_sample_207.png}
%     \caption{Uncurated generated samples of CelebA-HQ 512.}
%     \label{fig:uncurated_celeb512}
% \end{subfigure}%
% \begin{subfigure}{.45\textwidth}
%   \centering
%   \includegraphics[width=\linewidth]{figs/sit_sample_88.png}
%     \caption{Uncurated generated samples of CelebA-HQ 512.}
%     \label{fig:uncurated_celeb512}
% \end{subfigure}
% \end{figure}

\begin{figure}
    \begin{minipage}[c]{0.45\linewidth}
    \includegraphics[width=\linewidth]{figs/sit_207_sample_cfg4.0.png}
    \caption{Uncurated generated samples of ImageNet 256 class 207 (golden retriever) with cfg=4.0.}
    \label{fig:imgnetsupp_0}
    \end{minipage}
    \hfill
    \begin{minipage}[c]{0.45\linewidth}
    \includegraphics[width=\linewidth]{figs/sit_88_sample_cfg4.0.png}
    \caption{Uncurated generated samples of ImageNet 256 class 88 (macaw) with cfg=4.0.}
    \label{fig:imgnetsupp_1}
    \end{minipage}%
\end{figure}

% \begin{figure}
%     \includegraphics[width=0.45\linewidth]{figs/sit_sample_207.png}
%     \caption{Uncurated generated samples of CelebA-HQ 512.}
%     \label{fig:uncurated_celeb512}
% \end{figure}
% \hfill
% \begin{figure}
%     \includegraphics[width=0.45\linewidth]{figs/sit_sample_88.png}
%     \caption{Uncurated generated samples of CelebA-HQ 512.}
%     \label{fig:uncurated_celeb512}
% \end{figure}
\newpage
\section{NeurIPS Paper Checklist}
\begin{enumerate}

\item {\bf Claims}
    \item[] Question: Do the main claims made in the abstract and introduction accurately reflect the paper's contributions and scope?
    \item[] Answer: \answerYes{} % Replace by \answerYes{}, \answerNo{}, or \answerNA{}.
    \item[] Justification: Our contributions are accurately reflected in the abstract and in the summary of introduction section \cref{sec:intro}.

\item {\bf Limitations}
    \item[] Question: Does the paper discuss the limitations of the work performed by the authors?
    \item[] Answer: \answerYes{} % Replace by \answerYes{}, \answerNo{}, or \answerNA{}.
    \item[] Justification: We explicitly mention the limitation and suggest potential directions for future works in \cref{sec:conclusion}.

\item {\bf Theory Assumptions and Proofs}
    \item[] Question: For each theoretical result, does the paper provide the full set of assumptions and a complete (and correct) proof?
    \item[] Answer: \answerNA{} % Replace by \answerYes{}, \answerNo{}, or \answerNA{}.
    \item[] Justification: N/A, because our work focuses solely on empirical evaluation to measure the effectiveness of the proposed architecture.

    \item {\bf Experimental Result Reproducibility}
    \item[] Question: Does the paper fully disclose all the information needed to reproduce the main experimental results of the paper to the extent that it affects the main claims and/or conclusions of the paper (regardless of whether the code and data are provided or not)?
    \item[] Answer: \answerYes{} % Replace by \answerYes{}, \answerNo{}, or \answerNA{}.
    \item[] Justification: Yes, this information can be found in \cref{ap:training_details} and \cref{sec:exp}.

\item {\bf Open access to data and code}
    \item[] Question: Does the paper provide open access to the data and code, with sufficient instructions to faithfully reproduce the main experimental results, as described in supplemental material?
    \item[] Answer: \answerYes{} % Replace by \answerYes{}, \answerNo{}, or \answerNA{}.
    \item[] Justification: Yes, we will release our code and pretrained models on github.

\item {\bf Experimental Setting/Details}
    \item[] Question: Does the paper specify all the training and test details (e.g., data splits, hyperparameters, how they were chosen, type of optimizer, etc.) necessary to understand the results?
    \item[] Answer: \answerYes{} % Replace by \answerYes{}, \answerNo{}, or \answerNA{}.
    \item[] Justification: Yes, we provided them in Implementation part of \cref{sec:exp} and Training details in \cref{ap:training_details}.

\item {\bf Experiment Statistical Significance}
    \item[] Question: Does the paper report error bars suitably and correctly defined or other appropriate information about the statistical significance of the experiments?
    \item[] Answer: \answerNo{} % Replace by \answerYes{}, \answerNo{}, or \answerNA{}.
    \item[] Justification: No, because the evaluation process takes hours on a GPU and is very expensive to compute with multiple seeds. Importantly, we conducted evaluation for several benchmarks in \cref{sec:exp} so running with different seeds are not feasible in our scope.

\item {\bf Experiments Compute Resources}
    \item[] Question: For each experiment, does the paper provide sufficient information on the computer resources (type of compute workers, memory, time of execution) needed to reproduce the experiments?
    \item[] Answer: \answerYes{} % Replace by \answerYes{}, \answerNo{}, or \answerNA{}.
    \item[] Justification: Yes, this is presented in \cref{ap:training_details}.
    
\item {\bf Code Of Ethics}
    \item[] Question: Does the research conducted in the paper conform, in every respect, with the NeurIPS Code of Ethics \url{https://neurips.cc/public/EthicsGuidelines}?
    \item[] Answer: \answerYes{} % Replace by \answerYes{}, \answerNo{}, or \answerNA{}.
    \item[] Justification: Yes, we totally confirm to it.

\item {\bf Broader Impacts}
    \item[] Question: Does the paper discuss both potential positive societal impacts and negative societal impacts of the work performed?
    \item[] Answer: \answerYes{} % Replace by \answerYes{}, \answerNo{}, or \answerNA{}.
    \item[] Justification: We discussed the societal imparts in \cref{sec:conclusion}.
    
\item {\bf Safeguards}
    \item[] Question: Does the paper describe safeguards that have been put in place for responsible release of data or models that have a high risk for misuse (e.g., pretrained language models, image generators, or scraped datasets)?
    \item[] Answer: \answerYes{} % Replace by \answerYes{}, \answerNo{}, or \answerNA{}.
    \item[] Justification: Yes, our code will be released with Licences.

\item {\bf Licenses for existing assets}
    \item[] Question: Are the creators or original owners of assets (e.g., code, data, models), used in the paper, properly credited and are the license and terms of use explicitly mentioned and properly respected?
    \item[] Answer: \answerYes{} % Replace by \answerYes{}, \answerNo{}, or \answerNA{}.
    \item[] Justification: Yes, we have properly cited the code, data and models used.

\item {\bf New Assets}
    \item[] Question: Are new assets introduced in the paper well documented and is the documentation provided alongside the assets?
    \item[] Answer: \answerYes{} % Replace by \answerYes{}, \answerNo{}, or \answerNA{}.
    \item[] Justification: Our released source code is accompanied by an instruction document to run. Details about implementation can be found in section \cref{sec:exp}.

\item {\bf Crowdsourcing and Research with Human Subjects}
    \item[] Question: For crowdsourcing experiments and research with human subjects, does the paper include the full text of instructions given to participants and screenshots, if applicable, as well as details about compensation (if any)? 
    \item[] Answer: \answerNA{} % Replace by \answerYes{}, \answerNo{}, or \answerNA{}.
    \item[] Justification: The paper does not involve crowdsourcing nor research with human subjects.

\item {\bf Institutional Review Board (IRB) Approvals or Equivalent for Research with Human Subjects}
    \item[] Question: Does the paper describe potential risks incurred by study participants, whether such risks were disclosed to the subjects, and whether Institutional Review Board (IRB) approvals (or an equivalent approval/review based on the requirements of your country or institution) were obtained?
    \item[] Answer: \answerNA{} % Replace by \answerYes{}, \answerNo{}, or \answerNA{}.
    \item[] Justification: The paper does not involve crowdsourcing nor research with human subjects.
    
\end{enumerate}

\end{document}